\documentclass[11pt, a4paper]{mini}

\usepackage{lmodern} 
\usepackage{natbib}
\usepackage{subcaption}
\usepackage{multirow}
\usepackage{array}
\usepackage{float}
\usepackage{wrapfig}
\usepackage{needspace}
\usepackage[section]{placeins}
\usepackage{amsmath, amssymb}
\usepackage{capt-of}

\newcommand{\ghlogo}{\raisebox{-0.15em}{\includegraphics[height=1em]{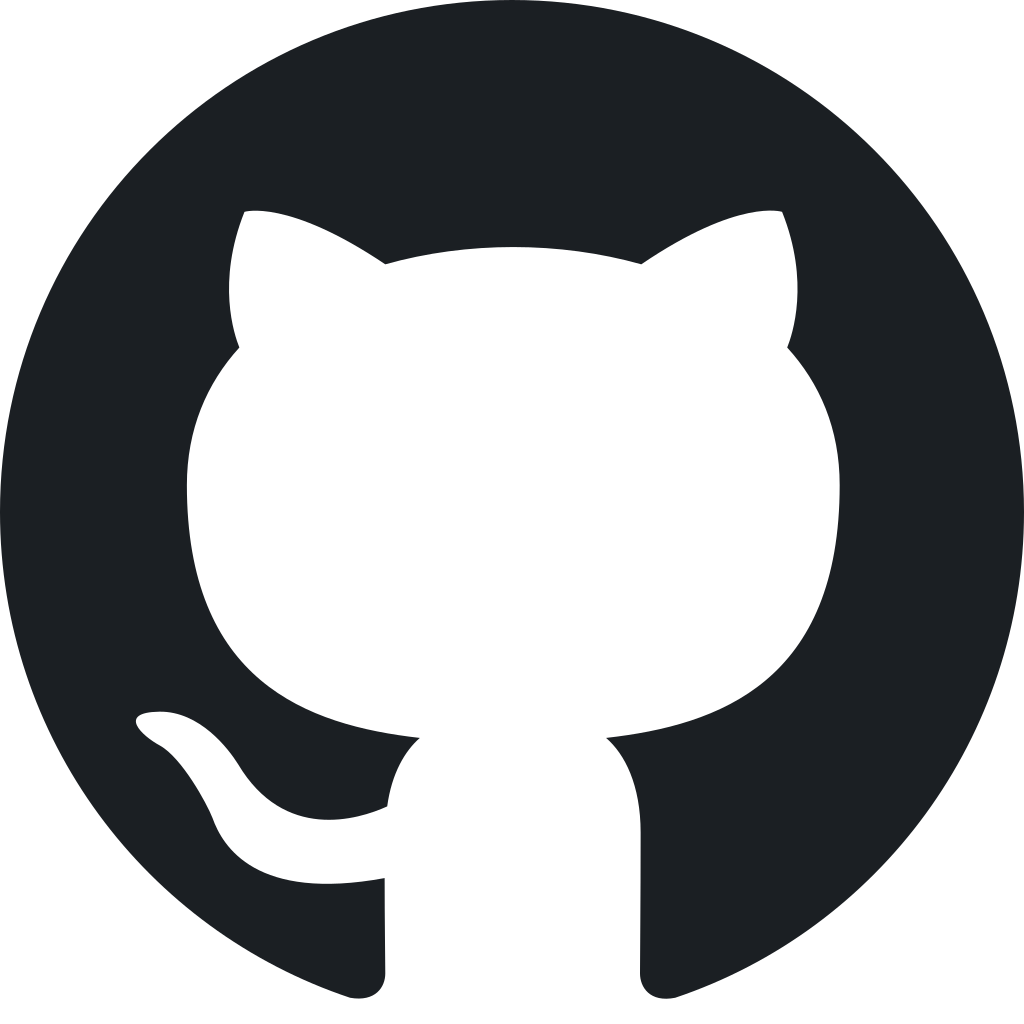}}}

\begin{document}

\title{Anatomically Guided Latent Diffusion for Brain MRI Progression Modeling}

\author[1,2]{Cheng Wan}
\author[2]{Bahram Jafrasteh}
\author[3]{Ehsan Adeli}
\author[4]{Miaomiao Zhang}
\author[2]{Qingyu Zhao}

\affil[1]{Cornell University}
\affil[2]{Weill Cornell Medicine}
\affil[3]{Stanford University}
\affil[4]{University of Virginia}

\correspondingauthor={Cheng Wan: cw2222@cornell.edu}

\maketitle

\begin{abstract}
Accurately modeling longitudinal brain MRI progression is crucial for understanding neurodegenerative diseases and predicting individualized structural changes. Existing state-of-the-art approaches, such as Brain Latent Progression (BrLP), often use multi-stage training pipelines with auxiliary conditioning modules but suffer from architectural complexity, suboptimal use of conditional clinical covariates, and limited guarantees of anatomical consistency. We propose Anatomically Guided Latent Diffusion Model (AG-LDM), a segmentation-guided framework that enforces anatomically consistent progression while substantially simplifying the training pipeline. AG-LDM conditions latent diffusion by directly fusing baseline anatomy, noisy follow-up states, and clinical covariates at the input level, a strategy that avoids auxiliary control networks by learning a unified, end-to-end model that represents both anatomy and progression.
A lightweight 3D tissue segmentation model (WarpSeg) provides explicit anatomical supervision during both autoencoder fine-tuning and diffusion model training, ensuring consistent brain tissue boundaries and morphometric fidelity. 
Experiments on 31,713 ADNI longitudinal pairs and zero-shot evaluation on OASIS-3 demonstrate that AG-LDM matches or surpasses more complex diffusion models, achieving highly competitive image quality and 15--20\% reduction in volumetric errors in generated images. AG-LDM also exhibits markedly stronger utilization of temporal and clinical covariates (3.5--31.5$\times$ higher covariate sensitivity than BrLP) and generates biologically plausible counterfactual trajectories, accurately capturing hallmarks of Alzheimer's progression such as limbic atrophy and ventricular expansion. These results highlight AG-LDM as an efficient, anatomically grounded framework for reliable brain MRI progression modeling.
\par\vspace{0.6em}\noindent
\ghlogo~\textbf{Code}: \href{https://github.com/JornyWan/AG-LDM}{\texttt{https://github.com/JornyWan/AG-LDM}}
\end{abstract}

\section{Introduction}
\label{sec:introduction}
\begin{figure*}[]    
\centering
  \includegraphics[width=\textwidth]{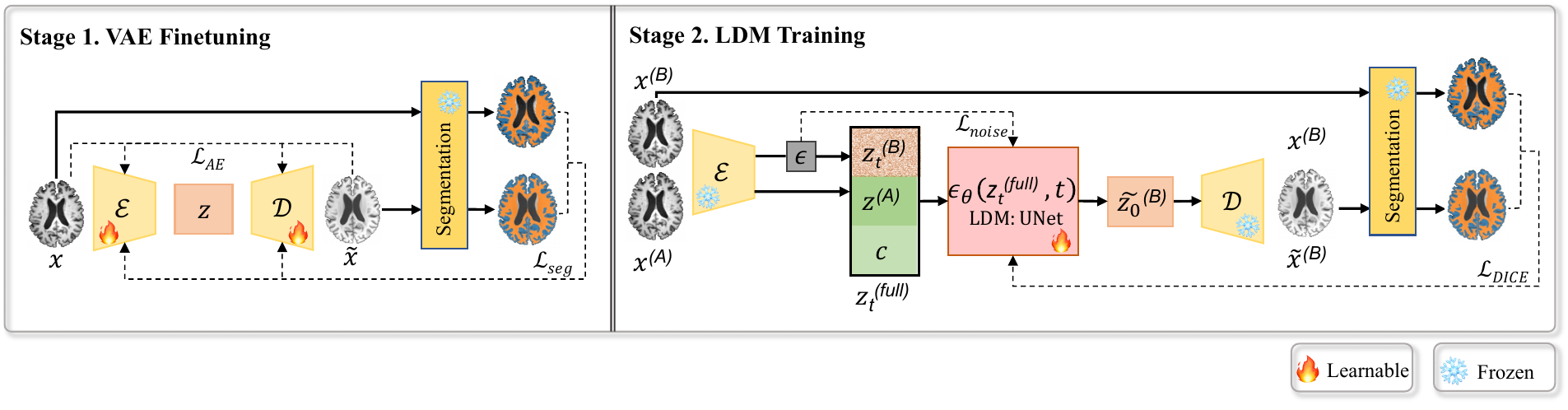}
  % \vspace{-0.8cm}
  \caption{AG-LDM is a two-stage brain MRI generative model: Stage 1 first fine-tunes an AE to learn latent representations of MRIs; In Stage 2, a channel-wise concatenation of the noisy follow-up latent $\mathbf{z}_t^{(B)}$, the latent of the baseline anatomy $\mathbf{z}^{(A)}$, and the clinical covariates $\mathbf{c}$ (e.g., sex, age and diagnosis information at time A and time B) is used as the input to train an LDM $\boldsymbol{\epsilon}_\theta$ that generates the latent of the future MRI. In both stages, segmentation guidance is used to ensure the reconstructed MRI has consistent gray matter (GM) and white matter (WM) tissue segmentation as the ground-truth.}
  \label{fig:overview}
\end{figure*}
The human brain undergoes profound, heterogeneous changes across the lifespan, with trajectories shaped by sociodemographic factors, life experiences, and diseases such as Alzheimer's disease (AD). The ability to model and predict individual-level structural brain changes is crucial for understanding how development and aging are influenced by external factors, and for enabling early diagnosis, treatment planning, and monitoring disease progression \cite{puglisi2024enhancing,young2024data}. %Accurate spatiotemporal modeling enables personalized medicine by providing patient-specific predictions that can inform clinical decision-making and improve outcomes.
While traditional statistical approaches \cite{oxtoby2017imaging,young2018uncovering} provide population-level insights, they fail to capture the complex spatiotemporal patterns characterizing individual disease trajectories.
To address this, deep generative models have emerged as a powerful paradigm to simulate future brain MRIs from past observations. Early approaches based on Generative Adversarial Networks (GANs) \cite{ravi2022degenerative,xia2021learning} and autoencoder (AE) models \cite{he2024individualized} often suffer from training instability or over-smoothing. More recently, diffusion models \cite{yoon2023sadm,litrico2024tadm,puglisi2024enhancing} have established a new standard, offering superior stability and high-quality generation.

Among recent developments, Brain Latent Progression (BrLP) \cite{puglisi2024enhancing} represents the current state-of-the-art for brain MRI progression modeling. BrLP combines latent diffusion models (LDMs) with a ControlNet module and an auxiliary disease progression module, achieving impressive performance across multiple datasets. 
However, BrLP's success comes at the cost of architectural complexity, requiring a three-stage training pipeline, in which the ControlNet module must learn to condition the diffusion process on baseline images through cross-attention mechanisms. The auxiliary disease progression module relies on having volume trajectories of five \textit{a-priori} selected brain regions specific to AD. This design increases computational overhead, restricts its applicability to one disease, and overlooks changes in other important brain regions.

Despite significant progress of BrLP in brain MRI progression modeling, several key issues remain unprobed. Traditional approaches train generative models by optimizing voxel-level intensity reconstruction and evaluate models by image-quality metrics. 
As synthetic MRIs achieve near-perfect visual quality, such image-quality metrics increasingly fail to capture clinically important morphometric errors \cite{peng2024metadata, jafrasteh2025wasabi}, necessitating training protocols that explicitly maintain accurate morphometric properties in regions and tissue types that are essential for clinical interpretation. 
Second, effectively incorporating conditional information remains challenging. While models can be conditioned on various clinical covariates (age, sex, diagnosis), ensuring that these conditions meaningfully influence the generation process is non-trivial. Multi-stage architectures with indirect conditioning mechanisms (e.g., through cross-attention or separate control modules) may not fully preserve individual-specific characteristics or leverage temporal information effectively, as conditioning signals must pass through multiple transformation stages during synthesis.

We propose Anatomically Guided Latent Diffusion Model (AG-LDM), a simplified framework for brain MRI progression modeling that addresses these challenges. Our approach combines a streamlined two-stage architecture with explicit anatomical supervision through segmentation guidance. Conditioned on a baseline image and clinical covariates, AG-LDM can generate a future MRI at a given age and clinical diagnosis while enforcing consistent tissue boundaries. AG-LDM achieves state-of-the-art performance while maintaining architectural simplicity and computational efficiency. Our key contributions are threefold:

1) Unified Progression Modeling: Instead of employing auxiliary control networks that inject conditioning as corrective guidance, AG-LDM treats baseline anatomy and clinical covariates as co-equal latent state variables fused directly into the backbone. This design models disease evolution as a unified conditional transition process, enabling stable end-to-end training without the need for multi-stage optimization or guidance-strength tuning.

2) Segmentation-Guided Training: We integrate a lightweight 3D brain tissue segmentation model (WarpSeg) to provide anatomical supervision during both AE fine-tuning and diffusion model training. Crucially, extending this supervision to the diffusion stage, not the AE stage alone, directly encourages the generated follow-up scans to preserve faithful tissue structure throughout the predicted progression.

3) Comprehensive Evaluation: Extensive experiments on two longitudinal brain MRI datasets demonstrate that AG-LDM achieves comparable or superior image quality and anatomical precision compared to state-of-the-art generative models. Our method reduces volumetric errors by 15--20\% in key anatomical regions and maintains strong generalization to unseen external datasets. Results also indicate that AG-LDM can better leverage conditional information in the generation process (e.g., 3.5--31.5$\times$ greater sensitivity to clinical covariates compared to baseline models). Counterfactual analysis reveals that AG-LDM correctly captures disease-specific progression patterns (e.g., limbic atrophy and ventricular expansion in AD).

\section{Related Work}
\label{sec:related}
\subsection{Diffusion Models in Medical Imaging}
Diffusion models have emerged as a powerful paradigm for high-quality image generation, offering superior training stability compared to GANs and better sample quality than AEs. The foundational Denoising Diffusion Probabilistic Model (DDPM) \cite{ho2020denoising} introduced a principled approach to generative modeling through iterative denoising. Given a data sample $\mathbf{z}_0$, DDPM defines a forward Markov process $q(\mathbf{z}_t | \mathbf{z}_0) = \mathcal{N}(\mathbf{z}_t; \sqrt{\bar{\alpha}_t} \mathbf{z}_0, (1-\bar{\alpha}_t)\mathbf{I})$ that progressively adds Gaussian noise over $T$ timesteps, where $\{\alpha_t\}_{t=1}^T$ is a predefined noise schedule with $\bar{\alpha}_t = \prod_{i=1}^t \alpha_i$. This allows efficient sampling of noisy versions $\mathbf{z}_t = \sqrt{\bar{\alpha}_t} \mathbf{z}_0 + \sqrt{1-\bar{\alpha}_t} \boldsymbol{\varepsilon}$ with $\boldsymbol{\varepsilon} \sim \mathcal{N}(0, \mathbf{I})$.
The model learns to reverse this process via a Markov chain $p_\theta(\mathbf{z}_{t-1}|\mathbf{z}_t) = \mathcal{N}(\mathbf{z}_{t-1}; \boldsymbol{\mu}_\theta(\mathbf{z}_t, t), \boldsymbol{\Sigma}_\theta(\mathbf{z}_t, t))$, where the mean $\boldsymbol{\mu}_\theta$ and variance $\boldsymbol{\Sigma}_\theta$ are parameterized by a neural network. 
In practice, the network predicts the noise $\boldsymbol{\epsilon}_\theta(\mathbf{z}_t, t)$, and the denoised mean is computed as:
\begin{equation}
\boldsymbol{\mu}_\theta(\mathbf{z}_t, t) = \frac{1}{\sqrt{\alpha_t}}\left(\mathbf{z}_t - \frac{1-\alpha_t}{\sqrt{1-\bar{\alpha}_t}}\boldsymbol{\epsilon}_\theta(\mathbf{z}_t, t)\right).
\label{eq:ddpm_mean}
\end{equation}
This reverse process enables image generation from pure Gaussian noise through iterative denoising.
In our implementation, following the standard DDPM~\cite{ho2020denoising} and LDM~\cite{rombach2022high} practice, the variance $\boldsymbol{\Sigma}_\theta(\mathbf{z}_t, t) = \sigma_t^2 \mathbf{I}$ is fixed to the forward-process variance schedule and is not learned; only the mean is parameterized via the neural noise predictor $\boldsymbol{\epsilon}_\theta$ (Eq.~\ref{eq:ddpm_mean}).

Building upon DDPM, LDMs \cite{rombach2022high} operate in a compressed latent space to significantly improve computational efficiency while maintaining high generation quality, often leveraging deterministic sampling strategies \cite{song2020denoising}. In the medical domain, such frameworks have demonstrated promising results for brain image synthesis \cite{pinaya2022brain} and disease progression modeling.

\subsection{Brain MRI Generative Models}
Deep generative models have been widely explored for brain MRI synthesis. These methods can be categorized based on their generation objective: unconditional/conditional synthesis of new subjects, and longitudinal progression modeling.
In the first category, the methods aim to generate high-quality MRIs from noise, often conditioned on metadata or clinical variables. Prior to diffusion models, GAN-based approaches like $\alpha$-WGAN~\cite{kwon2019generation} and CCE-GAN~\cite{xing2021cycle} are widely explored but often suffered from unstable training or mode collapse. The use of DDPM resolves this issue, but a direct training of diffusion models in the 3D voxel space, such as performed by 3D-DDPM~\cite{dorjsembe2022three}, incurs higher computational costs. In contrast, LDM~\cite{pinaya2022brain} operates in a compressed latent space, significantly improving computational efficiency. Another way to address memory constraints is to generate volumes via a slice-by-slice approach, such as MedGen3D~\cite{han2023medgen3d}, albeit with potential compromises in global 3D consistency. Building on these, BrainSynth~\cite{peng2024metadata} demonstrates high-quality synthesis through vector quantization and masked diffusion. Other approaches have explored conditional diffusion for slice-wise synthesis~\cite{peng2023generating} or integrated physical constraints into LDMs~\cite{lupke2024physics} to enhance the plausibility of multimodal synthesis. Synthetic MRIs from these models are often used for data augmentation in downstream tasks but rarely used for progression modeling.

To perform progression modeling, generative models have to simulate future MRIs conditioned on past data. To do so, 3D autoregressive diffusion~\cite{yoon2023sadm} and 2D slice-based diffusion models have been used for longitudinal generation.
Other architectures have also been explored; for instance, He et al.~\cite{he2024individualized} propose an AE-based prediction framework, NeuroAR~\cite{yesiloglu2025neural} utilizes autoregressive transformers to model subject-specific brain aging trajectories, and Xia et al.~\cite{xia2021learning} utilize GANs to simulate aging without longitudinal data.
Among biologically constrained approaches, 4D-DaniNet \cite{ravi2022degenerative} simulates progression using adversarial learning, while CounterSynth \cite{pombo2023equitable} employs a deformation-based strategy to warp the baseline scan to a target state. More recently, diffusion-based methods use target-conditioned latent-space manipulation to achieve \emph{counterfactual} trajectory synthesis~\cite{huang2025cycle,yeganeh2025latent}. Litrico et al.~\cite{litrico2025temporally} introduce a bidirectional temporal-consistency regularization, enforcing that forward and backward predictions along the time axis remain mutually consistent. Lozupone et al.~\cite{lozupone2026latent} propose a latent diffusion AE that learns a semantically meaningful latent space; although primarily designed for unsupervised representation learning, it also enables counterfactual trajectory traversal as a downstream property. % Such conditional, counterfactual generation, namely simulating individualized trajectories under contrastive clinical hypotheses, is a clinically meaningful use case distinct from predicting an unknown future label. 
AG-LDM is complementary to these works: rather than relying on temporal-consistency constraints or unsupervised latent-space traversal, it enforces spatial anatomical consistency through a frozen segmentation teacher under explicit clinical-covariate conditioning.
Lastly, BrLP \cite{puglisi2024enhancing} combines latent diffusion with ControlNet and an auxiliary disease progression model to achieve state-of-the-art performance, at the cost of a complex, multi-stage, disease-specific design.

\subsection{Anatomical Consistency in Medical Image Generation}
Ensuring anatomical consistency in generated medical images is crucial for clinical applications. 
Recent work has shown that conventional image-level metrics such as FID, MS-SSIM, and MMD, while useful for assessing perceptual quality, lack sensitivity to crucial morphometric fidelity. 
The WASABI (Wasserstein-Based Anatomical Brain Index) metric~\cite{jafrasteh2025wasabi} addresses this gap by explicitly measuring multivariate Wasserstein distance between segmentation-derived volumetric measures. Because it compares the joint distribution of these morphometric measures rather than pixel-level image quality, a scan with high peak signal-to-noise ratio (PSNR) can still score poorly on WASABI when its morphometry is implausible. Indeed, Jafrasteh et al. demonstrated that synthetic MRIs achieving near-perfect visual quality may still exhibit subtle anatomical inaccuracies that are difficult to detect by human experts or existing metrics, necessitating evaluation protocols that explicitly assess whether generated images reflect true brain anatomy.
Beyond evaluation, ensuring anatomical plausibility during training remains challenging. One promising direction is deformation-based synthesis~\cite{wang2025generating,wilms2022invertible,wu2025igg}. For example, MorphLDM~\cite{wang2025generating} generates novel images by applying synthesized deformation fields to a brain template to ensure morphological plausibility. More recently, Jayakumar et al.~\cite{jayakumar2026generative} model the spatiotemporal deformation of neurodegenerative brain anatomy with a 4D longitudinal diffusion model.
However, these methods assume that the generated images share an identical topological structure with a given template, up to diffeomorphism, which is an overly restricted assumption. 

%For our task of brain progression modeling, we require anatomical consistency between baseline and follow-up scans while capturing disease-related structural changes. 
The use of segmentation-guided training has shown particular promise in brain imaging applications. For instance, BrainSPADE \cite{fernandez2024generating} uses semantic masks as conditional inputs to control the layout and generate high-quality pathological brains. While effective for label-to-image synthesis where the target anatomy is predefined, such approaches inherently rely on the availability of ground-truth segmentation maps during inference, which restricts their direct applicability to disease progression modeling where the future anatomical state is unknown. To address this, our work employs a lightweight segmentation teacher (WarpSeg) to provide anatomical supervision via loss functions, enabling explicit enforcement of tissue boundary consistency during the progression prediction.
%%%%%% Methods %%%%%%
\section{Methods}
\label{sec:methods}
\subsection{Overview and Formulation}
We design AG-LDM as a two-stage framework for longitudinal brain MRI progression. Formally, let $\mathcal{X} \subset \mathbb{R}^{H \times W \times D}$ denote the space of 3D T1-weighted brain MRI volumes. For a subject, we observe a baseline scan $\mathbf{x}^{(A)} \in \mathcal{X}$ acquired at age $a^{(A)}$ and a follow-up scan $\mathbf{x}^{(B)} \in \mathcal{X}$ at age $a^{(B)} > a^{(A)}$. Associated with each scan is a set of clinical covariates $\mathbf{c} = \{a^{(A)}, a^{(B)}, \text{sex}, d^{(A)}, d^{(B)}\}$, where $d^{(A)}$ and $ d^{(B)}$ represent the categorical diagnostic status at baseline and follow-up, respectively. Our objective is to learn a conditional generative model $p_\theta(\mathbf{x}^{(B)} | \mathbf{x}^{(A)}, \mathbf{c})$ that can synthesize anatomically plausible follow-up scans given baseline observations and clinical context. In other words, with the target diagnostic state $d^{(B)}$ as an explicit input, AG-LDM is a \textit{conditional and counterfactual} trajectory generator: for a fixed baseline it can simulate how the same brain would evolve under different hypothetical clinical scenarios (e.g., sustained cognitive normality vs.\ conversion to AD). 

To achieve this while ensuring computational efficiency and stable training, we adopt a latent diffusion framework~\cite{rombach2022high} operating in a compressed latent space $\mathcal{Z} \subset \mathbb{R}^{3 \times H' \times W' \times D'}$. The pipeline comprises two stages: (i) learning an AE to map between a 3D scan and its latent representation, and (ii) learning a conditional LDM to generate the latent of the follow-up scan. Figure~\ref{fig:overview} illustrates the complete pipeline.

\subsection{Baseline Pre-trained Autoencoder}
\label{subsec:vae}
To learn a compact latent representation that preserves both low-level image fidelity and high-level anatomical structure, we adopt the 3D AE designed in \cite{pinaya2022brain} and initialize its parameters from their publicly available pre-trained model.%, which we subsequently fine-tune with anatomical supervision.

% %\subsubsection{Architecture}
% Specifically, the 3D variational autoencoder has an encoder $\mathcal{E}: \mathcal{X} \to \mathcal{Z}$ and a decoder $\mathcal{D}: \mathcal{Z} \to \mathcal{X}$. The encoder maps an input volume $\mathbf{x}$ to latent representation $\mathbf{z}$:
% \begin{equation}
% \mathbf{z} = \mathcal{E}(\mathbf{x}) = \boldsymbol{\mu}(\mathbf{x}) + \boldsymbol{\sigma}(\mathbf{x}) \odot \boldsymbol{\epsilon}, \quad \boldsymbol{\epsilon} \sim \mathcal{N}(0, \mathbf{I}),
% \label{eq:encoder}
% \end{equation}
% where $\boldsymbol{\mu}(\cdot)$ and $\boldsymbol{\sigma}(\cdot)$ are learnable functions parameterized by the encoder, and the latent dimension is $3$ channels at spatial resolution $(H', W', D')$. The decoder then reconstructs the input from the latent.
Specifically, the 3D AE has an encoder $\mathcal{E}: \mathcal{X} \to \mathcal{Z}$ and a decoder $\mathcal{D}: \mathcal{Z} \to \mathcal{X}$. The encoder maps an input volume $\mathbf{x}$ to a latent representation $\mathbf{z} = \mathcal{E}(\mathbf{x}) = \boldsymbol{\mu}(\mathbf{x}) + \boldsymbol{\sigma}(\mathbf{x}) \odot \boldsymbol{\varepsilon}$, where $\boldsymbol{\varepsilon} \sim \mathcal{N}(0, \mathbf{I})$ and functions $\boldsymbol{\mu}(\cdot)$ and $\boldsymbol{\sigma}(\cdot)$ are parameterized by the encoder. The latent dimension is $3$ channels at spatial resolution $(H', W', D')$. The decoder then reconstructs the input from the latent.

%\subsubsection{Training Objective}
This AE has been pre-trained on a large-scale brain MRI dataset \cite{pinaya2022brain} to minimize a composite loss that balances reconstruction fidelity, latent regularization, adversarial realism, perceptual quality, and feature-level consistency. The baseline AE objective is:
\begin{equation}
\begin{aligned}
\mathcal{L}_{\text{AE}} = \;&\mathbb{E}_{\mathbf{x} \sim p_{\text{data}}} \Big[ \underbrace{\|\mathbf{x} - \hat{\mathbf{x}}\|_1}_{\text{reconstruction}} + \underbrace{\beta_{\text{KL}} \cdot \text{KL}(q_\phi(\mathbf{z}|\mathbf{x}) \| p(\mathbf{z}))}_{\text{KL regularization}} \\
&+ \underbrace{\lambda_{\text{perc}} \cdot \mathcal{L}_{\text{perc}}(\mathbf{x}, \hat{\mathbf{x}})}_{\text{perceptual loss}} + \underbrace{\lambda_{\text{adv}} \cdot \mathcal{L}_{\text{adv}}(\hat{\mathbf{x}})}_{\text{adversarial loss}} + \underbrace{\lambda_{\text{fm}} \cdot \mathcal{L}_{\text{fm}}(\mathbf{x}, \hat{\mathbf{x}})}_{\text{feature matching}}],
\end{aligned}
\label{eq:ae_loss}
\end{equation}
where $\hat{\mathbf{x}} = \mathcal{D}(\mathcal{E}(\mathbf{x}))$ is the reconstruction. The perceptual loss uses a pre-trained 3D feature extractor capturing high-level semantic similarity. The adversarial loss uses a patch-based discriminator $\mathcal{D}_{\text{adv}}$ with spectral normalization to encourage realistic local texture. The feature matching loss $\mathcal{L}_{\text{fm}}$ stabilizes adversarial training by encouraging intermediate discriminator feature similarity between real and reconstructed images:
\begin{equation}
\mathcal{L}_{\text{fm}}(\mathbf{x}, \hat{\mathbf{x}}) = \frac{1}{L}\sum_{l=1}^{L} \|\mathcal{D}_{\text{adv}}^{(l)}(\mathbf{x}) - \mathcal{D}_{\text{adv}}^{(l)}(\hat{\mathbf{x}})\|_1,
\label{eq:fm_loss}  
\end{equation}
where $\{\mathcal{D}_{\text{adv}}^{(l)}\}_{l=1}^L$ denote the $L$ intermediate feature layers of the discriminator, and the discriminator features for real images are computed with gradients disabled.

\subsection{Stage 1: Fine-tuning with Anatomical Supervision}
\label{subsec:seg_framework}
The first stage of our model aims to fine-tune the pre-trained AE on all MRIs in the training data. A critical challenge in brain MRI progression modeling is maintaining anatomical consistency while capturing disease-related structural changes. We address this by integrating a lightweight tissue-segmentation teacher into the model such that the reconstructed scan and the input MRI have consistent tissue-segmentation results.

%\subsubsection{Segmentation Teacher}
\subsubsection{Segmentation Guidance}
To guide AE fine-tuning, we require a pre-trained segmentation teacher. Established tools are impractical here: SynthSeg~\cite{billot2023synthseg} is TensorFlow-based, incurring cross-framework overhead within our PyTorch pipeline, while FastSurfer~\cite{henschel2020fastsurfer} relies on slow 2D slice aggregation. Given the thousands of iterative calls during training, both are too inefficient for end-to-end optimization.

To resolve these challenges, we employ WarpSeg~\cite{warpseg_webpage}, an in-house-trained compact 3D U-Net-based segmentation network with 7.79M parameters that enables efficient repeated inference during training. WarpSeg is pre-trained on a large-scale brain segmentation dataset to generate 6-class tissue or region segmentation. 
%\subsubsection{Dice-based Anatomical Loss}
In our work, we initially focus on preserving WM and GM structures, as these two tissue types define the fundamental cortical and subcortical boundaries that should remain stable during disease progression, and their delineation is robust across scanners, protocols, and pathological states. %We do not consider CSF as a guidance because its delineation is sensitive to skull-stripping procedures.}
Specifically, we denote WarpSeg as a network $\Phi: \mathcal{X} \to \mathbb{R}^{6 \times H \times W \times D}$. For any input image $\mathbf{x}$, the segmentor output $\mathbf{s} = \Phi(\mathbf{x})$ contains softmax probability maps $\mathbf{s}_k \in [0,1]^{H \times W \times D}$, where $k=1$ corresponds to the label map for WM and $k=2$ for GM. For the reconstructed image $\hat{\mathbf{x}} = \mathcal{D}(\mathcal{E}(\mathbf{x}))$, its corresponding segmentation is $\hat{\mathbf{s}} = \Phi(\hat{\mathbf{x}})$. We define two complementary losses. First, the soft Dice loss measures region overlap:
\begin{equation}
\mathcal{L}_{\text{dice}}(\mathbf{x}, \hat{\mathbf{x}}) = 1 - \frac{1}{2}\sum_{k \in \{1,2\}} \frac{2 \langle \mathbf{s}_k, \hat{\mathbf{s}}_k \rangle + \delta}{\|\mathbf{s}_k\|_1 + \|\hat{\mathbf{s}}_k\|_1 + \delta},
\label{eq:dice_loss}
\end{equation}
where $\langle \cdot, \cdot \rangle$ denotes the inner product over all spatial locations, $\|\cdot\|_1$ denotes the $L_1$ norm, and $\delta = 10^{-7}$ is a small constant for numerical stability.
Second, to enforce sharp tissue boundaries, we define a cross-entropy based boundary loss:
\begin{equation}
\mathcal{L}_{\text{boundary}}(\mathbf{x}, \hat{\mathbf{x}}) = -\frac{1}{2}\sum_{k \in \{1,2\}} \sum_{i=1}^{HWD} s_{k,i} \log \hat{s}_{k,i}.
\label{eq:boundary_loss}
\end{equation}
The complete anatomical supervision for AE fine-tuning combines both terms:
\begin{equation}
\mathcal{L}_{\text{seg}}(\mathbf{x}, \hat{\mathbf{x}}) = \mathcal{L}_{\text{dice}}(\mathbf{x}, \hat{\mathbf{x}}) + \mathcal{L}_{\text{boundary}}(\mathbf{x}, \hat{\mathbf{x}}).
\label{eq:seg_loss}
\end{equation}

With such anatomical supervision, the AE is fine-tuned under the complete Stage-1 objective, which augments the baseline reconstruction loss (Eq.~\ref{eq:ae_loss}) with the segmentation supervision term $\mathcal{L}_{\text{seg}}$:
\begin{equation}
\mathcal{L}_{\text{AE-finetune}} = \mathcal{L}_{\text{AE}} + \lambda_{\text{seg}} \cdot \mathcal{L}_{\text{seg}}(\mathbf{x}, \hat{\mathbf{x}}),
\label{eq:ae_finetune}
\end{equation}
where $\lambda_{\text{seg}}$ is a hyperparameter.

Note, gradients flow only through the generated image $\hat{\mathbf{x}}$, while the segmentor $\Phi$  remains \emph{frozen} during training to provide consistent anatomical supervision without introducing additional trainable parameters. This practice ensures that WM and GM boundaries are faithfully preserved in the learned latent space.

\subsection{Stage 2: Conditional Latent Diffusion Model}
\label{subsec:ldm}
During Stage~2, the fine-tuned AE from Stage~1 is strictly frozen, operating as a deterministic mapping to a $3 \times H' \times W' \times D'$ latent space.
The goal of the second stage is now to model the conditional distribution of follow-up scans in the learned latent space; i.e., to generate the latent of the follow-up scan $\mathbf{z}^{(B)}$ conditioned on baseline information $\mathbf{z}^{(A)}$ and clinical covariates $\mathbf{c}$. Building upon the diffusion framework introduced in Sec.~\ref{sec:related}, we take the follow-up latent $\mathbf{z}_0^{(B)} = \mathcal{E}(\mathbf{x}^{(B)})$ and apply the forward process with noise $\boldsymbol{\varepsilon} \sim \mathcal{N}(0, \mathbf{I})$ to sample a noisy version $\mathbf{z}_t^{(B)}$ at any timestep $t$.
%This design treats longitudinal progression as a conditional transition within a unified latent space rather than separating the generative prior from the conditioning signal (as in ControlNet-style architectures).

\subsubsection{Unified Latent Space Fusion}
To implement this conditional transition in a unified latent space, we construct an augmented input at each diffusion timestep $t$ by channel-wise concatenation of the noisy follow-up latent $\mathbf{z}_t^{(B)}$, the clean baseline latent $\mathbf{z}^{(A)} = \mathcal{E}(\mathbf{x}^{(A)})$, and spatially broadcast clinical covariates $\mathbf{c} \in \mathbb{R}^{5 \times H' \times W' \times D'}$ (i.e., the 5-channel clinical vector $[a^{(A)}, a^{(B)}, \text{sex}, d^{(A)}, d^{(B)}]$ is replicated identically at every spatial location of the latent grid). This results in an 11-channel composite input (3 channels for noisy follow-up, 3 for baseline, 5 for covariates):
\begin{equation}
\mathbf{z}_t^{(\text{full})} = [\mathbf{z}_t^{(B)}, \mathbf{z}^{(A)}, \mathbf{c}] \in \mathbb{R}^{11 \times H' \times W' \times D'},
\label{eq:concat}
\end{equation}

% The conditioning variables are fused with the noisy follow-up latent by channel-wise concatenation: the noisy follow-up latent $\mathbf{z}_t^{(B)} \in \mathbb{R}^{3 \times H' \times W' \times D'}$, the clean baseline latent $\mathbf{z}^{(A)} \in \mathbb{R}^{3 \times H' \times W' \times D'}$, and the spatially broadcast covariate tensor $\mathbf{c} \in \mathbb{R}^{5 \times H' \times W' \times D'}$ are stacked along the channel axis into $\mathbf{z}_t^{(\text{full})} = [\mathbf{z}_t^{(B)}, \mathbf{z}^{(A)}, \mathbf{c}] \in \mathbb{R}^{11 \times H' \times W' \times D'}$ (Eq.~\ref{eq:concat}); the noise-prediction U-Net's input layer is correspondingly widened from 3 to 11 channels.

% Here the 5-dimensional clinical covariate vector $[a^{(A)}, a^{(B)}, \text{sex}, d^{(A)}, d^{(B)}]$ is spatially broadcast, i.e.\ its values are replicated identically at every voxel of the latent grid, to form the covariate tensor $\mathbf{c} \in \mathbb{R}^{5 \times H' \times W' \times D'}$.

By constructing this composite input, AG-LDM treats baseline anatomy, noisy follow-up states, and clinical covariates as co-equal latent variables. This allows a single network to jointly capture anatomy and progression, treating disease evolution as a conditional transition process without the need for auxiliary control networks or guidance-strength tuning. A noise prediction network $\boldsymbol{\epsilon}_\theta(\mathbf{z}_t^{(\text{full})}, t): \mathbb{R}^{11 \times H' \times W' \times D'} \times \mathbb{N} \to \mathcal{Z}$ processes this input to predict the noise component; through this objective the model learns the conditional distribution $p_\theta(\mathbf{x}^{(B)} \mid \mathbf{x}^{(A)}, \mathbf{c})$ of the follow-up scan.

The network $\boldsymbol{\epsilon}_\theta$ is implemented as a 3D U-Net following the architecture of LDM~\cite{pinaya2022brain} (detailed specifications are provided in Sec.~\ref{sec:experiments}). The primary objective minimizes the mean squared error (MSE) between predicted and true noise $\boldsymbol{\varepsilon}$, following standard diffusion model training:
\begin{equation}
\mathcal{L}_{\text{noise}} = \mathbb{E}_{t \sim \mathcal{U}(1,T), \boldsymbol{\varepsilon} \sim \mathcal{N}(0,\mathbf{I})} \big[\|\boldsymbol{\varepsilon} - \boldsymbol{\epsilon}_\theta(\mathbf{z}_t^{(\text{full})}, t)\|_2^2\big].
\label{eq:noise_loss}
\end{equation}
Finally, the denoised latent $\tilde{\mathbf{z}}_0^{(B)}$ is computed using the reverse process mean function defined in Eq.~\ref{eq:ddpm_mean}.

% All conditioning signals are fused exclusively at the U-Net input layer via channel-wise concatenation (Eq.~\ref{eq:concat}): baseline anatomy and noisy follow-up share the same VAE encoder, and the 5-channel covariate tensor is a spatial broadcast of the clinical vector. No auxiliary skip connections, side networks, or cross-attention are used at deeper resolutions; only the diffusion timestep $t$ is re-injected. A layer-level specification of the U-Net is given in Supplementary Table~S1.
% \vspace{0.4cm}
\subsubsection{Segmentation-Guided Training}
\begin{wrapfigure}{r}{0.55\textwidth}
\centering
\includegraphics[width=0.55\textwidth]{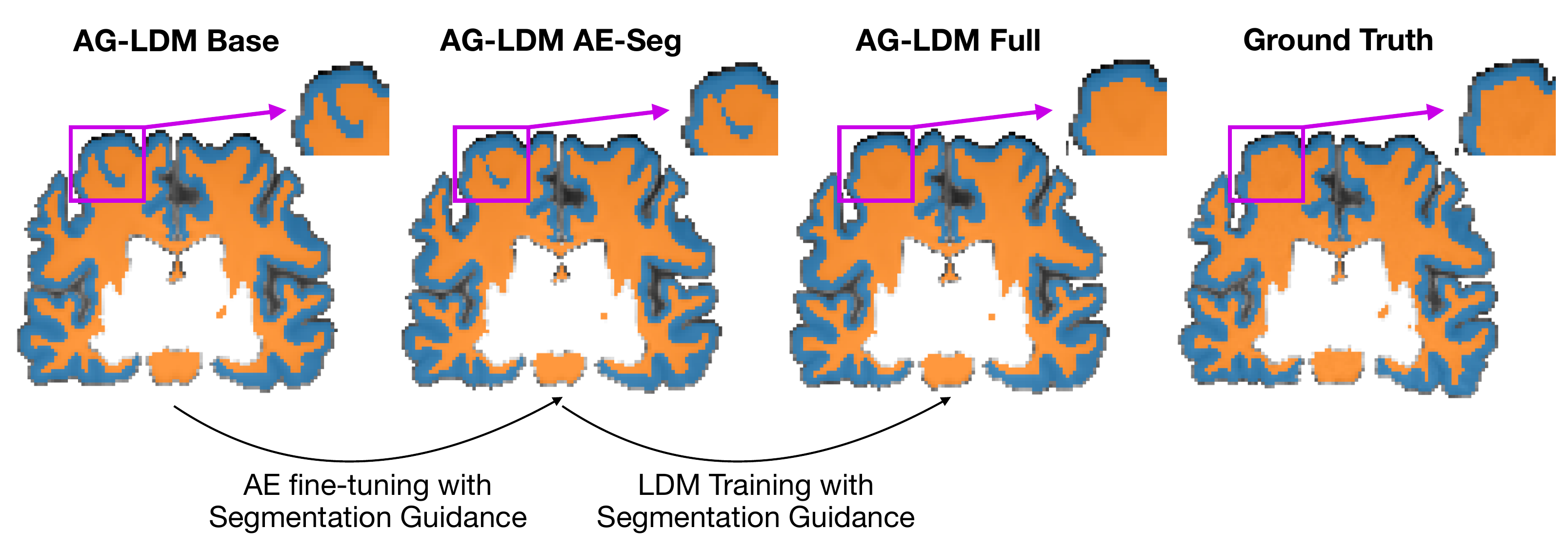}
\caption{Effect of segmentation-guided supervision: GM and WM segmentation of an MRI synthesized by AG-LDM Base, AG-LDM AE-Seg, and AG-LDM Full (conditioned on the same baseline MRI); Adding segmentation guidance at both stages produces segmentations closer to the ground truth.}
\label{fig:AG-LDM_variats_seg_visual}
\end{wrapfigure}
While the above noise prediction objective ensures accurate diffusion modeling, it does not explicitly enforce anatomical plausibility in generated images. To address this, we augment the training with anatomical supervision. At every training iteration we perform fast 10-step denoising via Denoising Diffusion Implicit Models (DDIM)~\cite{song2020denoising} from an intermediate noisy state to obtain a denoised latent $\tilde{\mathbf{z}}_0^{(B)}$. 
We then decode this to image space $\tilde{\mathbf{x}}^{(B)} = \mathcal{D}(\tilde{\mathbf{z}}_0^{(B)})$ to produce a generated follow-up scan and apply the Dice loss of Eq.~\ref{eq:dice_loss} between its tissue segmentation and that of the ground-truth follow-up. The complete Stage-2 training objective therefore augments the noise-prediction loss $\mathcal{L}_{\text{noise}}$ with this segmentation Dice term:
\begin{equation}
\mathcal{L}_{\text{LDM}} = \mathcal{L}_{\text{noise}} + \gamma \cdot \mathcal{L}_{\text{dice}}(\mathbf{x}^{(B)}, \tilde{\mathbf{x}}^{(B)}),
\label{eq:ldm_total}
\end{equation}
where $\gamma$ is a small weight balancing noise prediction accuracy with anatomical consistency. Throughout this supervision the AE ($\mathcal{E}, \mathcal{D}$) and the segmentation teacher $\Phi$ remain frozen, so $\boldsymbol{\epsilon}_\theta$ is the only updated module. Because $\mathcal{D}$ and $\Phi$ are differentiable, gradients of $\mathcal{L}_{\text{dice}}$ flow back through $\Phi\!\to\!\mathcal{D}\!\to\!\boldsymbol{\epsilon}_\theta$ across all 10 unrolled DDIM steps (no stop-gradient). 
As qualitatively demonstrated in Figure~\ref{fig:AG-LDM_variats_seg_visual}, the explicit anatomical supervision effectively suppresses spurious boundary artifacts, ensuring precise anatomical alignment with the ground truth.
\section{Experimental Setup}
\label{sec:experiments}
\subsection{Datasets and Preprocessing}
\subsubsection{ADNI Dataset}
% ================= Table I: Internal (ADNI) test set =================
\begin{table*}[!ht]
\caption{Quantitative comparison on the internal ADNI test set ($n=3{,}086$ baseline--follow-up pairs).
Image fidelity (MSE, PSNR), anatomy-aware structural consistency (WASABI, $\times 10^{-4}$), and Brain-Region Volume Mean Absolute Error (MAE, \%) for four regions of interest (ROIs): Amyg. (Amygdala), Hippo. (Hippocampus),
Lat. Vent. (Lateral Ventricles), Thal. (Thalamus). Per-sample metrics are mean $\pm$ standard deviation (SD); best results are in \textbf{bold}; $^{*}$ marks the best entry when it significantly outperforms every other method (paired one-sided $t$-test, $p<0.001$). WASABI is a cohort-level score and is not tested.}
\vspace{-0.6em}
\label{tab:comparison_adni}
\centering
\small
\setlength{\tabcolsep}{4pt}
\renewcommand{\arraystretch}{1.2}
\resizebox{\textwidth}{!}{
\begin{tabular}{p{3.1cm}|cc|c|cccc}
\toprule
\multirow{2}{*}{\textbf{Method}}
& \multicolumn{2}{c|}{\textbf{Image Fidelity}}
& \textbf{Anatomical}
& \multicolumn{4}{c}{\textbf{Brain-Region Volume MAE (\%)} $\downarrow$} \\
\cmidrule(lr){2-3}\cmidrule(lr){4-4}\cmidrule(lr){5-8}
& MSE $\downarrow$ & PSNR $\uparrow$ & WASABI $\downarrow$
& Amyg. & Hippo. & Lat.\ Vent. & Thal. \\
\midrule
4D-DaniNet~\cite{ravi2022degenerative}
& $0.027 \pm 0.014$ & $15.84 \pm 2.10$ & $14.83$
& $0.055 \pm 0.039$ & $0.118 \pm 0.086$ & $1.750 \pm 1.610$ & $0.168 \pm 0.116$ \\

Latent\text{-}SADM~\cite{yoon2023sadm}
& $0.022 \pm 0.007$ & $16.59 \pm 1.72$ & $2.59$
& $0.030 \pm 0.028$ & $0.073 \pm 0.058$ & $1.434 \pm 1.401$ & $0.117 \pm 0.086$ \\

CounterSynth~\cite{pombo2023equitable}
& $0.005 \pm 0.004$ & $23.19 \pm 2.54$ & $3.66$
& $0.016 \pm 0.010$ & $0.028 \pm 0.023$ & $0.308 \pm 0.374$ & $0.039 \pm 0.028$ \\

BrLP~\cite{puglisi2024enhancing}
& $0.006 \pm 0.004$ & $22.16 \pm 2.14$ & $8.25$
& $0.028 \pm 0.023$ & $0.044 \pm 0.048$ & $\mathbf{0.190 \pm 0.606}^{*}$ & $\mathbf{0.035 \pm 0.051}^{*}$ \\
\midrule
\makebox[\linewidth][r]{AG-LDM}
& $\mathbf{0.003 \pm 0.002}^{*}$ & $\mathbf{24.74 \pm 2.08}^{*}$ & $\mathbf{0.44}$
& $\mathbf{0.013 \pm 0.011}^{*}$ & $\mathbf{0.026 \pm 0.021}^{*}$ & $0.293 \pm 0.250$ & $0.037 \pm 0.028$ \\
\bottomrule
\end{tabular}}
\end{table*}

% ================= Table II: External (OASIS-3) test set =================
\begin{table*}[!ht]
\caption{Quantitative comparison on the external OASIS-3 test set ($n=1{,}965$ baseline--follow-up pairs).
Per-sample metrics are reported as mean $\pm$ SD across the test-set pairs; 
$^{*}$ marks the best entry when it significantly outperforms every other method (paired
one-sided $t$-test, $p<0.001$). WASABI is a cohort-level score and is not tested.}
\vspace{-0.6em}
\label{tab:comparison_oasis}
\centering
\small
\setlength{\tabcolsep}{4pt}
\renewcommand{\arraystretch}{1.2}
\resizebox{\textwidth}{!}{
\begin{tabular}{p{3.1cm}|cc|c|cccc}
\toprule
\multirow{2}{*}{\textbf{Method}}
& \multicolumn{2}{c|}{\textbf{Image Fidelity}}
& \textbf{Anatomical}
& \multicolumn{4}{c}{\textbf{Brain-Region Volume MAE (\%)} $\downarrow$} \\
\cmidrule(lr){2-3}\cmidrule(lr){4-4}\cmidrule(lr){5-8}
& MSE $\downarrow$ & PSNR $\uparrow$ & WASABI $\downarrow$
& Amyg. & Hippo. & Lat.\ Vent. & Thal. \\
\midrule
4D-DaniNet~\cite{ravi2022degenerative}
& $0.025 \pm 0.016$ & $15.62 \pm 2.20$ & $9.17$
& $0.027 \pm 0.022$ & $0.095 \pm 0.080$ & $1.690 \pm 1.606$ & $0.140 \pm 0.115$ \\

Latent\text{-}SADM~\cite{yoon2023sadm}
& $0.023 \pm 0.007$ & $16.34 \pm 1.92$ & $4.84$
& $0.029 \pm 0.026$ & $0.065 \pm 0.055$ & $1.308 \pm 1.172$ & $0.089 \pm 0.081$ \\

CounterSynth~\cite{pombo2023equitable}
& $0.004 \pm 0.004$ & $24.28 \pm 2.93$ & $1.83$
& $0.013 \pm 0.013$ & $0.024 \pm 0.025$ & $0.380 \pm 0.417$ & $0.049 \pm 0.043$ \\

BrLP~\cite{puglisi2024enhancing}
& $0.005 \pm 0.003$ & $22.91 \pm 2.48$ & $7.31$
& $0.031 \pm 0.018$ & $0.060 \pm 0.042$ & $0.773 \pm 0.586$ & $0.056 \pm 0.044$ \\
\midrule
\makebox[\linewidth][r]{AG-LDM}
& $\mathbf{0.003 \pm 0.001}^{*}$ & $\mathbf{25.03 \pm 1.32}^{*}$ & $\mathbf{0.70}$
& $\mathbf{0.011 \pm 0.009}^{*}$ & $\mathbf{0.021 \pm 0.018}^{*}$ & $\mathbf{0.264 \pm 0.294}^{*}$ & $\mathbf{0.041 \pm 0.031}^{*}$ \\
\bottomrule
\end{tabular}}
\end{table*}

We use 10,678 T1-weighted MRI scans from 1,920 subjects in the Alzheimer's Disease Neuroimaging Initiative (ADNI)~\cite{petersen2010alzheimer} dataset\footnote{\href{http://adni.loni.usc.edu/}{http://adni.loni.usc.edu/}}, with 56.2\% female, baseline ages of 54.9--94.6 years (mean: 74.3 years), and baseline diagnoses of 34.9\% cognitively normal (CN), 55.1\% mild cognitive impairment (MCI), and 10.1\% AD.
%Each subject underwent multiple longitudinal scans over follow-up periods up to 15.92 years. 
To construct longitudinal pairs, we exhaustively pair all temporally ordered scans within each subject.
This yields 31,713 longitudinal pairs with a mean interval of 2.89 years (median: 2.01 years, maximum: 15.92 years). We perform subject-level splitting in an 8:1:1 ratio: 25,498 pairs (1,534 subjects) for training, 3,129 pairs (194 subjects) for validation, and 3,086 pairs (192 subjects) for testing. No subject appears across multiple splits, eliminating subject-level leakage between training and evaluation.

\subsubsection{OASIS-3 External Validation}
For external validation, we use the Open Access Series of Imaging Studies (OASIS-3)~\cite{lamontagne2019oasis}\footnote{\href{https://www.oasis-brains.org/}{https://www.oasis-brains.org/}} dataset. The dataset comprises 573 subjects (41.5\% female) with baseline ages of 43.0--92.0 years (mean: 66.9 years) and a bimodal diagnostic distribution (76.6\% CN, 2.2\% MCI, 21.2\% AD). Following the same pairing strategy, we exhaustively pair all temporally ordered scans within each subject, yielding 1,965 pairs with mean scan interval of 4.57 years. After training on ADNI, the model is directly tested on OASIS-3 without fine-tuning, providing zero-shot generalization evaluation.

\subsubsection{Preprocessing}
All scans undergo standardized preprocessing: N4 bias-field correction \cite{tustison2010n4itk}, skull stripping \cite{hoopes2022synthstrip}, affine registration to MNI152 space, and resampling to 1.5mm isotropic resolution. Images are padded/cropped to $122 \times 146 \times 122$ voxels, with intensities clipped at 5th--99.5th percentiles and min-max normalized to $[0,1]$.

\subsection{Model Architecture}
\label{sec:impl_details}
\label{sec:vis10y}
\subsubsection{Autoencoder} 
\begin{figure*}[!ht]
\centering
\includegraphics[width=\textwidth]{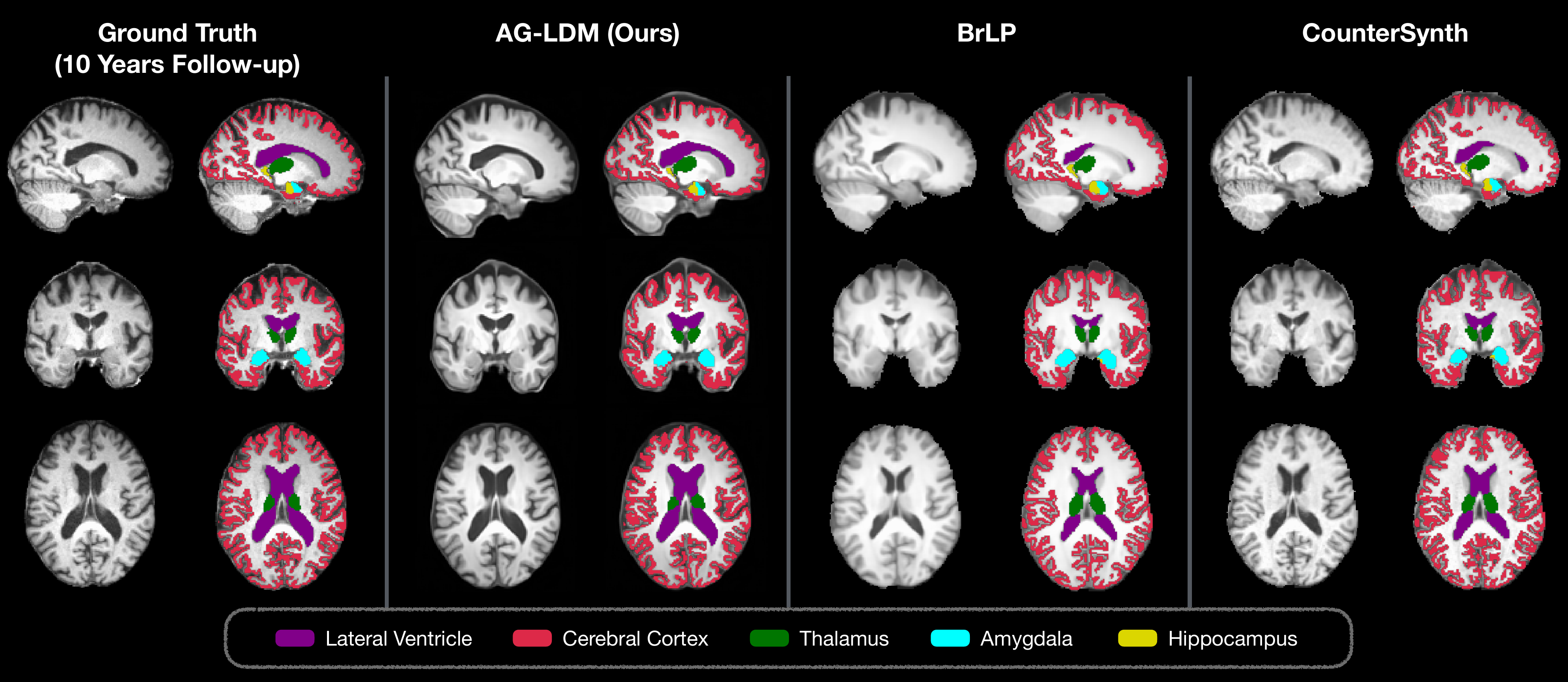}
\caption{\textbf{Ten-year brain progression synthesis (73$\rightarrow$83 years) for a randomly selected AD patient:} Columns correspond to the ground truth follow-up scan and scans generated by AG-LDM, BrLP, and CounterSynth. For each method we show matched sagittal/coronal/axial slices with SynthSeg overlays (right image in each pair). All four columns correspond to the same subject and follow-up scan. Identical sagittal/coronal/axial slices with SynthSeg overlays (right image in each pair) are visualized for direct structural comparison.}
\label{fig:vis10yr_montage}
\end{figure*}

\begin{wraptable}{l}{0.55\textwidth}
\vspace{-0.5em}
\centering
\small
\renewcommand{\arraystretch}{1.15}
\caption{Computational comparison of BrLP and AG-LDM on the same NVIDIA H100 GPU and ADNI split (25{,}498 training pairs); inference latency and peak VRAM are averaged over 5 runs at 50 DDIM steps.}
\label{tab:complexity}
\resizebox{\linewidth}{!}{%
\begin{tabular}{@{}l c c@{}}
\toprule
\textbf{Component / Metric} & \textbf{BrLP} & \textbf{AG-LDM (Ours)} \\
\midrule
Autoencoder (frozen)            & $13.76$~M & $13.76$~M \\
Diffusion U-Net (trainable)     & $553.20$~M & $\mathbf{475.44}$~M \\
ControlNet (trainable)          & $229.66$~M & --- \\
Segmentation teacher (frozen)   & --- & $7.79$~M \\
\textbf{Total trainable params} & $782.86$~M & $\mathbf{475.44}$~M \\
\midrule
Training stages                 & $3$ & $\mathbf{2}$ \\
Total training time (h)         & $\sim$$13.2$ & $\sim$$26.0$ \\
\midrule
Inference time / sample (s)     & $2.46$ & $\mathbf{1.01}$ \\
Peak VRAM (GB)                  & $15.02$ & $\mathbf{11.27}$ \\
\bottomrule
\end{tabular}}
\vspace{-0.4cm}
\end{wraptable}

We adopt the 3D AE architecture from the MONAI Generative Models library \cite{pinaya2022brain}, following the same configuration as BrLP \cite{puglisi2024enhancing}. The AE uses a 3-channel latent representation with an encoder that progressively downsamples through four levels with channel dimensions $(64, 128, 128, 128)$, using 2 residual blocks per level with group normalization (32 groups). The latent space achieves approximately $8\times$ spatial downsampling. We initialize from publicly available pre-trained weights\footnote{Pretrained AE is available at: \href{https://github.com/Project-MONAI/GenerativeModels/tree/main/model-zoo/models/brain_image_synthesis_latent_diffusion_model}{https://github.com/Project-MONAI/GenerativeModels}} and fine-tune with anatomical supervision.

\subsubsection{Segmentation Teacher} For anatomical supervision, we use an in-house 3D segmentation model based on the WarpSeg architecture\footnote{Our WarpSeg is available at: \href{https://github.com/BahramJafrasteh/WarpSeg}{https://github.com/BahramJafrasteh/WarpSeg}}. WarpSeg is a semi-supervised, dual-decoder deep learning framework designed for data-efficient 3D brain MRI segmentation. The architecture features a shared encoder that extracts multi-scale features, feeding into two distinct decoders: a coarse decoder that predicts six major tissue classes (background, GM, WM, ventricles, CSF, and deep gray matter) and a fine decoder that segments 30 detailed anatomical regions. 
Only the output for the GM and WM is used for training. WarpSeg is pre-trained on a large-scale brain segmentation dataset and remains frozen during all training stages.
\begin{figure*}[!t]
\centering
\includegraphics[width=\textwidth]{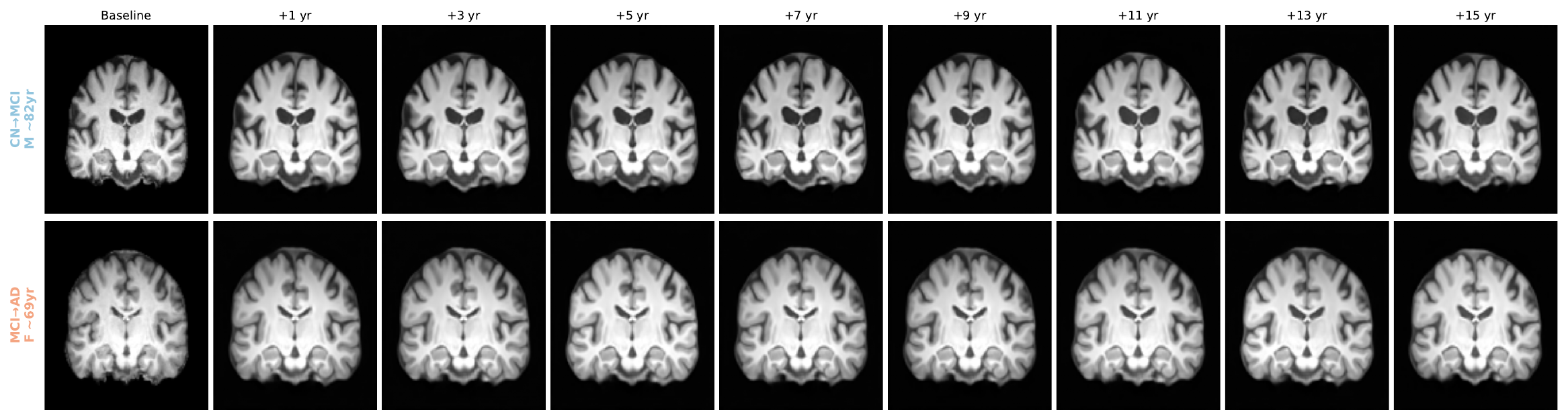}
\caption{Continuous-time trajectory generation for two held-out ADNI test subjects covering clinical transitions (top: CN$\rightarrow$MCI; bottom: MCI$\rightarrow$AD). Each row shows the baseline scan (leftmost) followed by AG-LDM predictions generated at uniform 1-year intervals from the same baseline scan (every other year, $+1,+3,\ldots,+15$), varying only the target age $a^{(B)}$ (and $d^{(B)}$). AG-LDM produces smooth trajectories consistent with the underlying clinical state: mild atrophy onset in CN$\rightarrow$MCI and pronounced ventricular expansion with cortical thinning in MCI$\rightarrow$AD.}
% \vspace{-2mm}
\label{fig:multisub_traj}
\end{figure*}

\subsubsection{Diffusion Model} The noise prediction network $\boldsymbol{\epsilon}_\theta$ adopts the MONAI 3D U-Net architecture \cite{pinaya2022brain}, consistent with BrLP's implementation \cite{puglisi2024enhancing}. The network comprises three downsampling levels with channel dimensions $(256, 512, 768)$. Self-attention mechanisms are applied at the two deepest resolutions using single-layer transformers. As described in Sec.~\ref{subsec:ldm}, the network processes 11 input channels through channel concatenation (Eq.~\ref{eq:concat}). Each level uses 2 residual blocks with group normalization (32 groups).

\subsection{Training and Inference}

\subsubsection{Training Configuration}
Based on the performance on the ADNI validation set, we select optimal hyperparameters and training configurations for our model and comparison methods. Given that AG-LDM is designed from a simplified version of BrLP, we ensure that the overlapping hyperparameters between BrLP and AG-LDM are chosen to be the same.
Specifically, in the first stage as Sec.~\ref{subsec:seg_framework}, we fine-tune the pretrained AE for 10 epochs using Adam optimizer with learning rate $\eta_{\text{AE}} = 5 \times 10^{-6}$ and effective batch size 16 (achieved via gradient accumulation from physical batch size 2). Adjusting from the hyperparameter setting in BrLP, we set KL divergence weight $\beta_{\text{KL}} = 1 \times 10^{-6}$, perceptual loss weight $\lambda_{\text{perc}} = 0.08$, adversarial loss weight $\lambda_{\text{adv}} = 0.1$, feature matching weight $\lambda_{\text{fm}} = 10$, and anatomical segmentation weight $\lambda_{\text{seg}} = 1.0$. The discriminator uses spectral normalization for training stability. We employ mixed precision training (FP16) with automatic mixed precision (AMP) to reduce memory consumption. This serves as the primary training configuration for our model in all experiments.

Following standard LDM training protocols \cite{rombach2022high}, we apply a global latent space normalization to ensure unit variance. We compute a scale factor $s = 1/\text{Std}(\mathbf{z})$ from 1,000 randomly sampled training latents, scaling inputs by $s$ before diffusion training and rescaling by $1/s$ during decoding.

\begin{wrapfigure}{r}{0.5\textwidth}
% \vspace{-1.0em}
\centering
\includegraphics[width=0.48\textwidth]{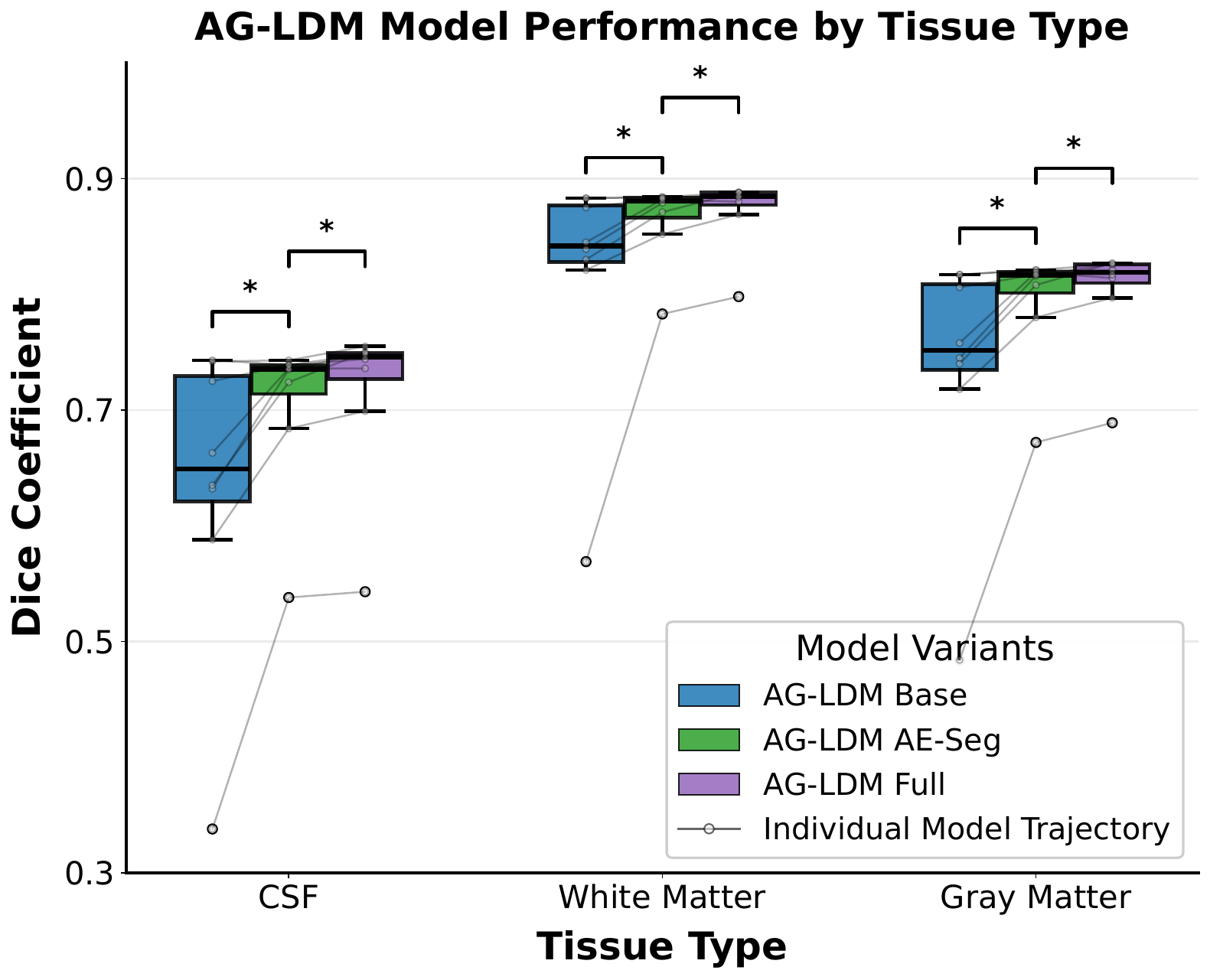}
\vspace{-0.3em}
\caption{Anatomical consistency measured by Dice coefficient for GM, WM, and CSF on the validation set. Each box represents 8 AG-LDM variants trained with different hyperparameter configurations. Brackets denote significant improvement ($^{*}p<0.001$) of each consecutive variant (Base, AE-Seg, and Full) measured by paired one-sided $t$-tests.}
\label{fig:four_AG-LDM_models}
\vspace{-0.8cm}
\end{wrapfigure}
In the second stage, we train the conditional LDM for 20 epochs using AdamW optimizer with learning rate $\eta_{\text{LDM}} = 2.5 \times 10^{-5}$ and batch size 8. The diffusion model follows DDPM \cite{ho2020denoising} with $T = 1000$ training timesteps and a scaled linear noise schedule with $\beta_1 = 0.0015$ and $\beta_T = 0.0205$. Following Eq.~\ref{eq:ldm_total}, the segmentation Dice supervision (Sec.~\ref{subsec:ldm}) is applied at every iteration with Dice loss weight $\gamma=1\times10^{-5}$; the small $\gamma$ keeps the segmentation gradient about an order of magnitude below the noise-prediction gradient, so this dense supervision does not destabilize training.

% The Dice loss weight $\gamma$ was selected on the ADNI validation set. A sensitivity analysis over four orders of magnitude $\gamma \in [10^{-7}, 10^{-3}]$ (Figure~\ref{fig:gamma_sens}; Supplementary Table~S2 gives the underlying numerical values) shows that the method is essentially insensitive to $\gamma$ within $[10^{-7}, 10^{-4}]$, with $\gamma=10^{-5}$ achieving both the lowest MSE and the highest WM/GM Dice.

The 20-epoch schedule corresponds to approximately 63{,}750 optimizer steps ($25{,}498$ training pairs at batch size 8). Empirical loss tracking shows that validation loss plateaus within the first $\sim$16{,}000 steps ($\sim$5 epochs) and remains flat for the remaining $\sim$47{,}800 steps, confirming full convergence and no overfitting; extending the training budget further yields no additional gain.

All experiments are conducted on a single NVIDIA H100 GPU (96GB). Training the AE takes approximately 8.5~hours, and training the diffusion model takes approximately 17.5~hours. Inference takes about 1~second per sample.

\subsubsection{Inference and Evaluation}
\begin{table*}[t]
\caption{Ablation study on Internal (ADNI) and External (OASIS-3) test sets. 
Mean and SD are computed over 8 hyperparameter configurations for each AG-LDM variant.}
\label{tab:comparison_volmae_ablation}
\centering
\small
\setlength{\tabcolsep}{4pt}
\renewcommand{\arraystretch}{1.1}
\resizebox{\textwidth}{!}{
\begin{tabular}{p{3.1cm}|cccc|cccc}
\toprule
\multirow{2}{*}{\textbf{Method}}
& \multicolumn{4}{c|}{\textbf{Internal (ADNI) -- Volume MAE (\%)} $\downarrow$} 
& \multicolumn{4}{c}{\textbf{External (OASIS-3) -- Volume MAE (\%)} $\downarrow$} \\
\cmidrule(lr){2-5}\cmidrule(lr){6-9}
& Amyg. & Hippo. & Lat.\ Vent. & Thal.
& Amyg. & Hippo. & Lat.\ Vent. & Thal. \\
\midrule
\textbf{AG-LDM\hfill Base}
& $0.042 \pm 0.064$ & $0.105 \pm 0.152$ & $0.763 \pm 0.904$ & $0.136 \pm 0.266$
& $0.040 \pm 0.069$ & $0.103 \pm 0.168$ & $0.635 \pm 0.756$ & $0.139 \pm 0.279$ \\

\makebox[\linewidth][r]{\textbf{AE-Seg}}
& $0.025 \pm 0.018$ & $0.067 \pm 0.052$ & $0.452 \pm 0.144$ & $\textbf{0.076} \pm \textbf{0.052}$
& $0.017 \pm 0.009$ & $\textbf{0.050} \pm \textbf{0.040}$ & $0.368 \pm 0.114$ & $\textbf{0.084} \pm \textbf{0.056}$ \\

\makebox[\linewidth][r]{\textbf{Full}}
& $\textbf{0.022} \pm \textbf{0.012}$ & $\textbf{0.057} \pm \textbf{0.050}$ & $\textbf{0.411} \pm \textbf{0.235}$ & $0.082 \pm 0.051$
& $\textbf{0.016} \pm \textbf{0.005}$ & $0.053 \pm 0.047$ & $\textbf{0.353} \pm \textbf{0.179}$ & $0.087 \pm 0.054$ \\

\bottomrule
\end{tabular}}
\end{table*}
For any given scan pair from the ADNI testing split or the entire OASIS-3, the model generates the follow-up scan conditioned on the baseline scan and clinical covariates.
We use DDIM \cite{song2020denoising} with 50 inference steps for efficient generation at test time. %the model iteratively denoises conditioned on the baseline latent and clinical covariates.

The generated follow-up scan is compared with the ground-truth follow-up scan by MSE. Quality of the generated scans is measured by PSNR. Beyond image quality, we also measure the anatomical accuracy of the generated scans. To avoid potential bias introduced by WarpSeg, we segment both the generated scan and ground truth using an independent segmentation tool SynthSeg \cite{billot2023synthseg} (FreeSurfer 7.4.1). Then we report tissue-level WM/GM/CSF Dice between the generated and ground-truth segmentations. We also compute the volume for four clinically relevant ROIs that are also used for evaluation in BrLP: amygdala, hippocampus, lateral ventricles, and thalamus.

We normalize these regional volumes by head size and calculate the absolute percentage error between predicted and ground-truth volumes, i.e. Brain-Region Volume MAE (\%). Lastly, given that most SynthSeg-derived ROI volumes are approximately Gaussian (with only mild skewness in the lateral ventricles), we report the WASABI score to assess multivariate Wasserstein distance of the 18 regional volume measures between generated and ground truth scans. 

The training and inference code for AG-LDM, including the Stage-1 fine-tuning utilities, Stage-2 segmentation-guided diffusion training routines, configurations for all ablation variants, inference scripts, and pretrained checkpoints, is publicly available at \url{https://github.com/JornyWan/AG-LDM}.

\section{Results}
\subsection{Comparison with State-of-the-Art Methods}
We compare AG-LDM against four state-of-the-art methods for longitudinal brain MRI progression modeling: 4D-DaniNet \cite{ravi2022degenerative}, Latent-SADM \cite{yoon2023sadm}, CounterSynth \cite{pombo2023equitable}, and BrLP \cite{puglisi2024enhancing}. Notably, BrLP serves as the representative state-of-the-art baseline for LDMs equipped with ControlNet conditioning and auxiliary progression modules. Tables~\ref{tab:comparison_adni} and~\ref{tab:comparison_oasis} summarize image quality and anatomical precision for four selected ROIs on the ADNI test set and the OASIS-3 external validation set, respectively, with all per-sample metrics reported as mean $\pm$ SD across the test-set pairs.

On ADNI, AG-LDM achieves the best performance on five of the seven metrics: the best image quality (MSE, PSNR), lowest WASABI, and lowest volumetric MAE on Amygdala and Hippocampus (paired one-sided $t$-test, $p<0.001$). BrLP attains the lowest Lateral-Ventricle and Thalamus MAE, where it significantly outperforms AG-LDM ($p<0.001$), likely because its auxiliary module explicitly optimizes these regional volumes; however, BrLP suffers from a substantially higher WASABI score, indicating that it preserves specific regional volumes at the cost of global morphometric plausibility.

Figure~\ref{fig:vis10yr_montage} shows an example of the generated scan at 10-year follow-up for a randomly selected AD patient. Because AG-LDM conditions on continuous chronological age rather than a discrete interval label, it can synthesize intermediate progression states at any target age; Figure~\ref{fig:multisub_traj} shows continuous 15-year trajectories for representative held-out subjects. While all methods generate visually appealing MRIs, AG-LDM generates brain structures more closely aligned with the ground truth. In contrast, BrLP and CounterSynth exhibit noticeably smaller lateral ventricles and enlarged thalamus regions in these slices, whereas AG-LDM accurately preserves the morphometry of these structures. 
%Because this is an extreme long-horizon case, the absolute errors here are larger than the test-set means reported in Table~\ref{tab:comparison_adni}.

On the external OASIS-3 test set, AG-LDM achieves the best performance on all seven metrics. Paired one-sided $t$-tests confirm that AG-LDM significantly outperforms every baseline on every per-sample metric ($p<0.001$). Against BrLP in particular, AG-LDM achieves 2--3$\times$ lower errors on all four ROIs and substantially lower WASABI, indicating stronger generalization than BrLP on the external dataset.

Beyond accuracy, AG-LDM is also computationally lighter than BrLP (Table~\ref{tab:complexity}): it uses 39\% fewer trainable parameters (475.44~M vs.\ 782.86~M), runs $2.4\times$ faster at inference (1.01~s vs.\ 2.46~s), and uses 25\% less GPU memory (11.27 vs.\ 15.02~GB); its longer total training time ($\sim$26 vs.\ $\sim$13~h) reflects the additional differentiable anatomical supervision and does not affect inference speed.

% \vspace{1.0em}
\subsection{Ablation Studies}
Next, we inspect the influence of adding segmentation guidance to both stages of our model. We evaluate three distinct scenarios to isolate the impact of our contributions: removing segmentation supervision from both training stages (Base), which functions as a standard conditional LDM baseline using the same channel-wise conditioning but no anatomical supervision; applying segmentation guidance only during

\newpage

\begin{wraptable}{r}{0.55\textwidth}
% \vspace{-1.0em}
\centering
\footnotesize
\setlength{\tabcolsep}{3.2pt}
\renewcommand{\arraystretch}{1.12}
% \vspace{-0.3em}
\caption{AE reconstruction quality of Base and AE-Seg across 8 hyperparameter configurations. For each variant we compute the average Dice or MAE over the ADNI test data and report mean $\pm$ SD of that metric over the 8 configurations. $^{*}$ AE-Seg significantly outperforms Base (paired one-sided $t$-test  $p<0.001$). Best in bold.}
\label{tab:ae_recon}
\resizebox{0.55\textwidth}{!}{%
\begin{tabular}{l c c}
\toprule
\textbf{Metric} & \textbf{Base ($\lambda_{\text{seg}}{=}0$)} & \textbf{AE-Seg ($\lambda_{\text{seg}}{=}1$)} \\
\midrule
GM Dice $\uparrow$        & $0.812 \pm 0.050$ & $\mathbf{0.874 \pm 0.011}^{*}$ \\
WM Dice $\uparrow$        & $0.879 \pm 0.034$ & $\mathbf{0.918 \pm 0.007}^{*}$ \\
CSF Dice $\uparrow$       & $0.766 \pm 0.049$ & $\mathbf{0.853 \pm 0.015}^{*}$ \\
\midrule
GM Vol-MAE (\%) $\downarrow$  & $3.39 \pm 1.72$  & $\mathbf{1.15 \pm 0.44}^{*}$ \\
WM Vol-MAE (\%) $\downarrow$  & $4.14 \pm 1.51$  & $\mathbf{0.88 \pm 0.19}^{*}$ \\
CSF Vol-MAE (\%) $\downarrow$ & $18.50 \pm 7.62$ & $\mathbf{4.27 \pm 2.88}^{*}$ \\
\bottomrule
\end{tabular}}
% \vspace{-0.4cm}
\end{wraptable}

\noindent
AE fine-tuning (AE-Seg); and incorporating anatomical supervision across both AE and diffusion training stages (Full). To ensure our finding is not biased by the specific hyperparameter setting derived from the ADNI validation set,  we train the AE under 8 different hyperparameter configurations by varying learning rates ($\eta_{\text{AE}} \in \{10^{-7}$ to $5{\times}10^{-5}\}$), feature matching weights ($\lambda_{\text{fm}} \in \{0, 1, 10\}$), perceptual loss weights ($\lambda_{\text{perc}} \in \{0.005$ to $0.08\}$), and KL divergence weights ($\beta_{\text{KL}} \in \{10^{-7}$ to $5{\times}10^{-5}\}$). This yields 24 models (8 configurations $\times$ 3 variants).
Crucially, all models use identical Stage 2 hyperparameters and global latent normalization, isolating the effect of anatomical supervision applied during AE fine-tuning.

Figure~\ref{fig:four_AG-LDM_models} presents tissue-level anatomical consistency measured by Dice coefficient for two primary tissue types (WM and GM) and CSF across the entire brain, and Table~\ref{tab:comparison_volmae_ablation} records volume MAE for the 4 ROIs similar to Tables~\ref{tab:comparison_adni} and~\ref{tab:comparison_oasis}.
Results indicate progressive benefits of anatomical supervision.
\begin{wrapfigure}{r}{0.55\textwidth}
% \vspace{-1.0em}
\centering
\includegraphics[width=0.55\textwidth]{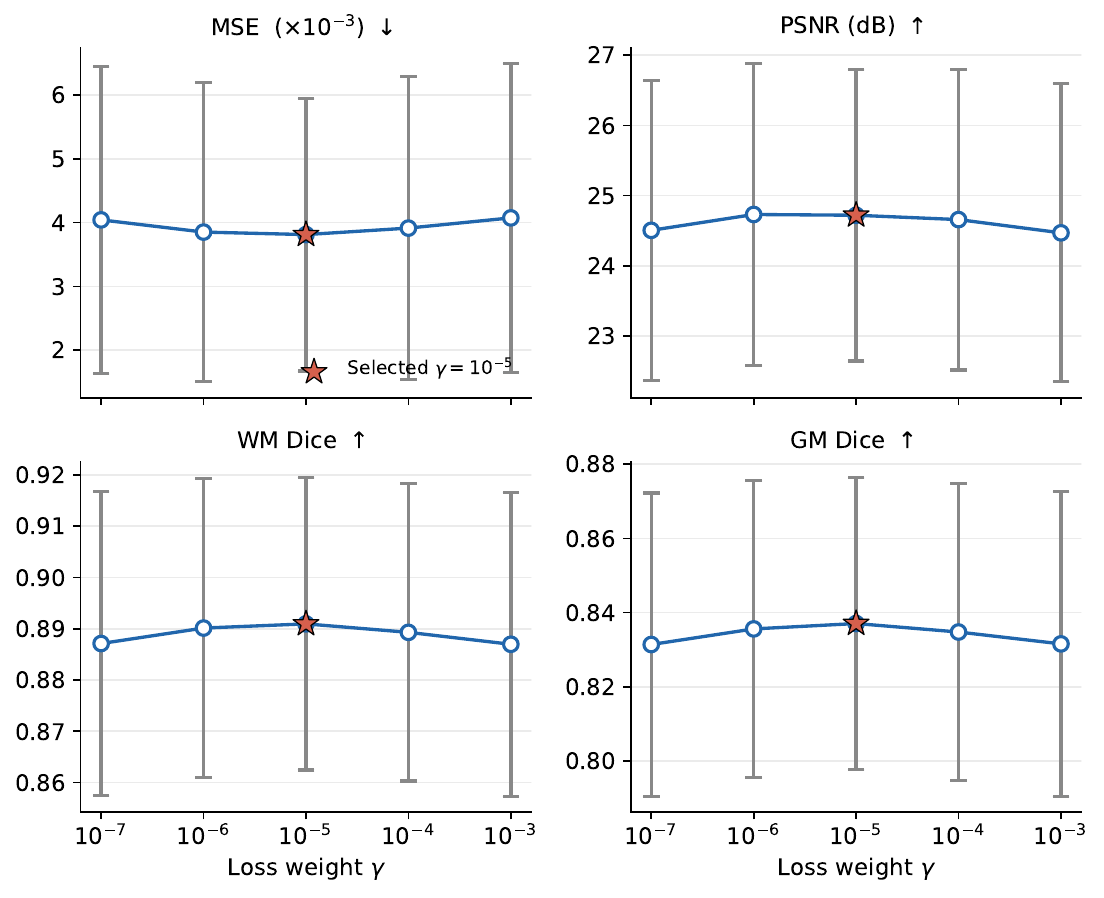}
\footnotesize
% \vspace{-0.3em}
\caption{Sensitivity of AG-LDM to the segmentation-loss weight $\gamma$ on the ADNI validation set. Markers show the mean across validation pairs and error bars are $\pm$1~SD. The red star marks the selected value $\gamma=10^{-5}$. Performance is essentially flat across three orders of magnitude $\gamma \in [10^{-7},10^{-4}]$, confirming insensitivity to this hyperparameter.}
\label{fig:gamma_sens}
\vspace{-0.4cm} 
\end{wrapfigure}
Starting from \textbf{Base}, adding segmentation guidance during AE fine-tuning (\textbf{AE-Seg}) significantly improves tissue- and ROI-level anatomical consistency, particularly for CSF. In addition to assessing the quality of generated follow-up scans, we further compare the self-reconstruction quality of Base ($\lambda_{\text{seg}}=0$) and AE-Seg ($\lambda_{\text{seg}}=1$) on the baseline scans of the ADNI test set, repeated across the same 8 matched hyperparameter configurations. Table~\ref{tab:ae_recon} reports tissue-level Dice and Volume MAE between the input scan and its reconstruction; AE-Seg significantly improves every anatomical metric (paired one-sided $t$-tests $p<0.001$).

% \vspace{10pt}
Next, incorporating anatomical supervision during diffusion training (\textbf{Full}) yields moderate yet statistically significant additional gains over AE-Seg for WM and GM preservation (paired one-sided $t$-test over the 8 validation configurations, $p<0.001$ for GM, WM, and CSF; Figure~\ref{fig:four_AG-LDM_models}). The narrower boxplot distributions associated with the AG-LDM variants with segmentation guidance compared to the base model demonstrate that segmentation guidance not only improves performance but also stabilizes training, making the approach robust to hyperparameter choices. To show this, a sensitivity analysis over the loss weight $\gamma$ (Figure~\ref{fig:gamma_sens}) shows that the method is robust over three orders of magnitude, with the chosen $\gamma=1\times10^{-5}$ showing the optimal performance.

Finally, to justify the unified channel-concatenation fusion adopted in Sec.~\ref{subsec:ldm}, we compare it against two standard conditioning mechanisms (Classifier-Free Guidance and Cross-Attention) under matched architectures and training budgets (Table~\ref{tab:cond_ablation}). Channel concatenation achieves the lowest MSE, the highest PSNR, and the highest mean ROI Dice (0.842 across 18 SynthSeg regions), and is statistically tied with Cross-Attention on global anatomical plausibility (WASABI, $1.13$ vs.\ $1.08$) and on regional volume MAE ($0.093$ vs.\ $0.090\%$). CFG is consistently the weakest of the three. Channel concatenation moreover avoids the $\sim$78M extra trainable parameters required by a Cross-Attention re-implementation of $\boldsymbol{\epsilon}_\theta$ and the per-step unconditional-pass sampling required by CFG. For the task of conditional longitudinal synthesis, treating baseline anatomy and covariates as co-equal input channels is therefore at least as effective as the more complex conditioning pathways while being substantially simpler.

\subsection{Conditioning Effectiveness Analysis}
\begin{wraptable}{r}{0.55\textwidth}
% \vspace{-1.0em}
\centering
% \footnotesize
\renewcommand{\arraystretch}{1.15}
\vspace{-0.3em}
\caption{Ablation of conditioning strategies for $\boldsymbol{\epsilon}_\theta$ on the ADNI validation split. All variants share identical AE checkpoint, U-Net backbone, and training schedule. Per-sample metrics are reported as mean $\pm$ SD across the validation-set pairs; WASABI ($\times 10^{-4}$) is a cohort-level point estimate. Best in bold.}
\label{tab:cond_ablation}
\resizebox{0.55\textwidth}{!}{%
\begin{tabular}{@{}l ccc@{}}
\toprule
\textbf{Metric} & \textbf{Channel-Concat (ours)} & \textbf{CFG} & \textbf{Cross-Attention} \\
\midrule
MSE ($\times 10^{-3}$) $\downarrow$  & $\mathbf{3.812 \pm 2.143}$ & $4.609 \pm 3.110$ & $4.207 \pm 2.478$ \\
PSNR (dB) $\uparrow$                  & $\mathbf{24.720 \pm 2.080}$ & $24.062 \pm 2.370$ & $24.312 \pm 2.091$ \\
WASABI $\downarrow$                   & $1.13$ & $1.34$ & $\mathbf{1.08}$ \\
Avg ROI Dice $\uparrow$               & $\mathbf{0.842 \pm 0.026}$ & $0.828 \pm 0.036$ & $0.838 \pm 0.027$ \\
Avg ROI MAE (\%) $\downarrow$         & $0.093 \pm 0.066$ & $0.102 \pm 0.074$ & $\mathbf{0.090 \pm 0.075}$ \\
\bottomrule
\end{tabular}}
\vspace{-0.4cm}
\end{wraptable}
Next, we evaluate how effectively our model utilizes conditional clinical covariates for brain MRI progression synthesis. The idea is that if the model does not learn conditional information, the conditional distribution degenerates to the marginal distribution, resulting in generated scans that are independent of the clinical covariates.

We randomly sample 50 subjects from the ADNI test set. For each subject, we generate follow-up predictions under different conditioning scenarios by selectively removing individual conditional inputs. When a condition is removed, we replace it with a neutral default value: continuous variables (starting age, follow-up age) are set to the dataset median, while categorical variables (sex, cognitive diagnosis) are set to zero.

All other conditions remain at their true values. We then measure the MSE between the generated and real follow-up images. The sensitivity to each condition is quantified as the percentage change in MSE relative to the model using full clinical covariates: $\text{Sensitivity} = (\text{MSE}_{\text{removed}} - \text{MSE}_{\text{full}}) / \text{MSE}_{\text{full}} \times 100\%$, where larger positive values indicate the model is more sensitive to the conditional covariates.

\begin{wraptable}{r}{0.55\textwidth}
\vspace{-0.3em}
\caption{Conditional Input Sensitivity Analysis: Percentage change (Mean $\pm$ SD) in generation error (MSE) after replacing each conditional covariate with a neutral value. Larger positive values indicate greater sensitivity to conditional variables.}
\label{tab:conditioning_sensitivity}
\centering
\renewcommand{\arraystretch}{1.1}
\setlength{\tabcolsep}{3pt}
\resizebox{0.55\textwidth}{!}{%
\begin{tabular}{l|r|r|c}
\toprule
\textbf{Removed Variable} &
\multicolumn{1}{c|}{\textbf{AG-LDM (Ours)}} &
\multicolumn{1}{c|}{\textbf{BrLP}} &
\textbf{Fold Diff.} \\
\midrule
Starting age        & $+408.3 \pm 397.4\%$ & $+16.4 \pm 42.8\%$ & \textbf{24.9$\times$} \\
Follow-up age       & $+138.4 \pm 189.1\%$ & $-4.4 \pm 38.0\%$  & \textbf{31.5$\times$} \\
Sex                 & $+43.6 \pm 149.7\%$  & $+11.9 \pm 51.8\%$ & \textbf{3.6$\times$}  \\
Cognitive diagnosis & $+36.2 \pm 112.6\%$  & $+10.2 \pm 50.1\%$ & \textbf{3.5$\times$}  \\
\bottomrule
\end{tabular}}
% \vspace{-0.4cm}
\end{wraptable}
Given that BrLP represents the current state-of-the-art for brain MRI progression modeling, also supported by our previous experiments, and is the only publicly available baseline with comparable conditioning capabilities, we compare AG-LDM's conditioning effectiveness against BrLP to quantify the practical benefits of our simplified architecture.

The results reveal a clear contrast in conditioning effectiveness between the two architectures (Table~\ref{tab:conditioning_sensitivity}). AG-LDM exhibits substantially higher sensitivity to conditional inputs, demonstrating 3.5--31.5$\times$ stronger reliance on clinical covariates than BrLP. Notably, omitting the starting age condition in AG-LDM results in a 408\% increase in MSE, and omitting follow-up age results in a 138\% increase, while BrLP exhibits only 16\% MSE change with respect to starting age and 4.4\% MSE change to follow-up age. 
For sex and diagnosis information, AG-LDM demonstrates moderate but significant sensitivity (44\% and 36\% MSE change), while BrLP's output remains largely unaffected (12\% and 10\%). A paired two-sided $t$-test across the 50 subjects confirms that AG-LDM's sensitivity to every covariate is significantly greater than BrLP's ($p<0.001$ for starting age, follow-up age, sex, and cognitive diagnosis). These results confirm that direct channel-wise concatenation at the input level of LDM may be a more effective conditioning strategy than a segregated ControlNet.

\subsection{Counterfactual Study}
\label{sec:counterfactual}

\newcommand{\sixw}{0.155\textwidth}

\begin{figure*}[!t]
\centering
% ---------- Row 1: CN -> AD ----------
\begin{subfigure}[t]{\sixw}\centering
  \includegraphics[width=\linewidth]{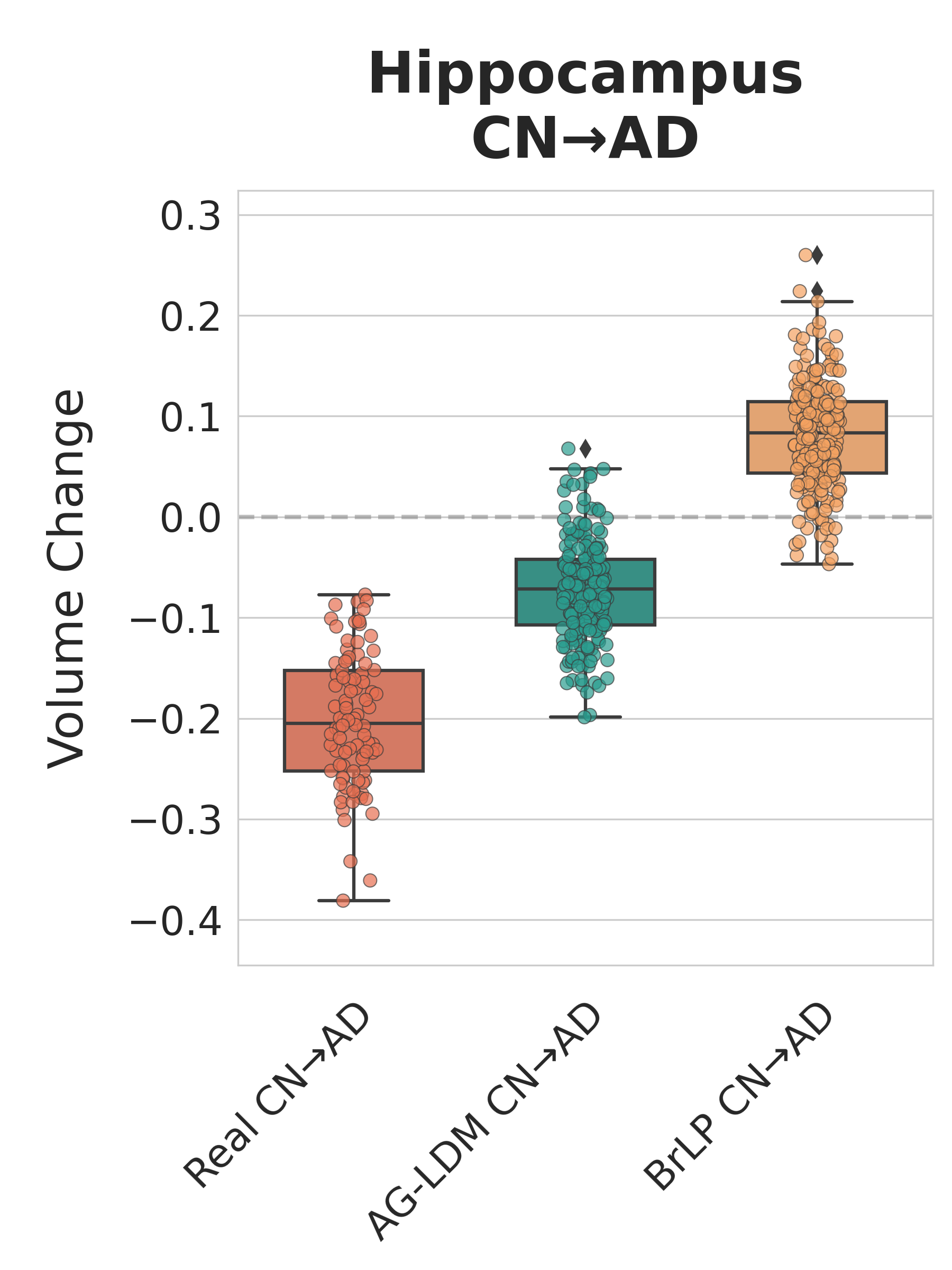}
\end{subfigure}\hfill
\begin{subfigure}[t]{\sixw}\centering
  \includegraphics[width=\linewidth]{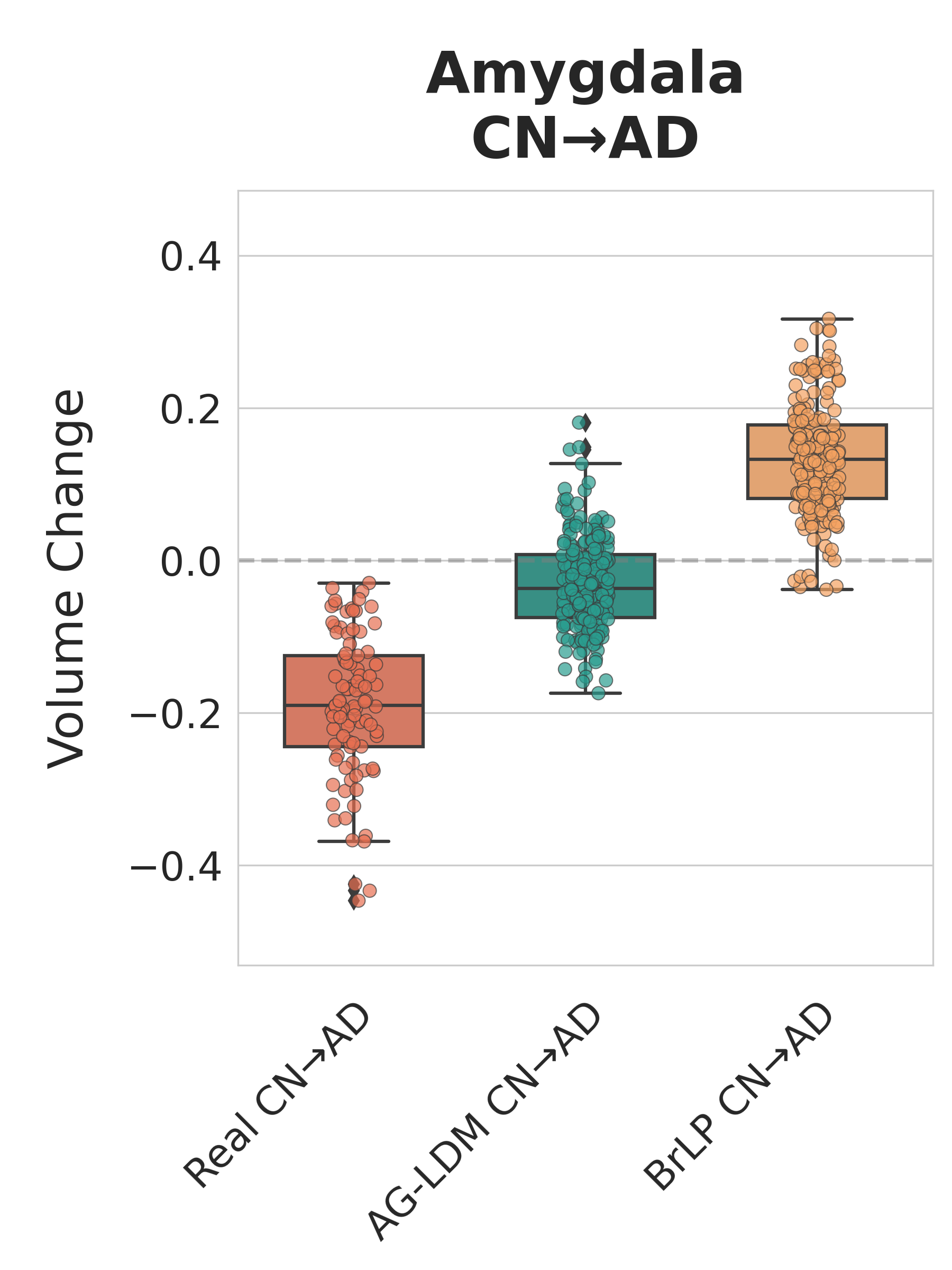}
\end{subfigure}\hfill
\begin{subfigure}[t]{\sixw}\centering
  \includegraphics[width=\linewidth]{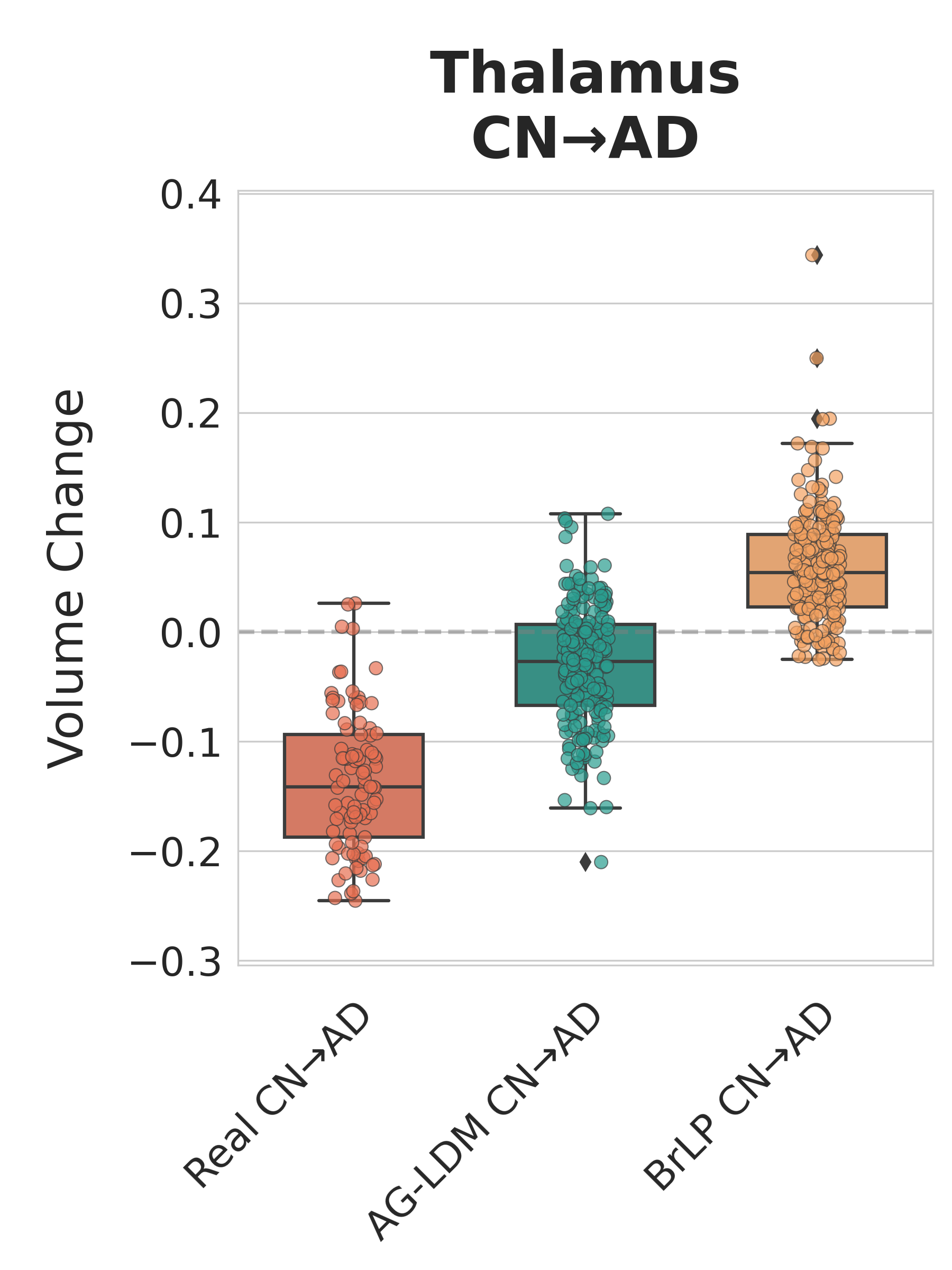}
\end{subfigure}\hfill
\begin{subfigure}[t]{\sixw}\centering
  \includegraphics[width=\linewidth]{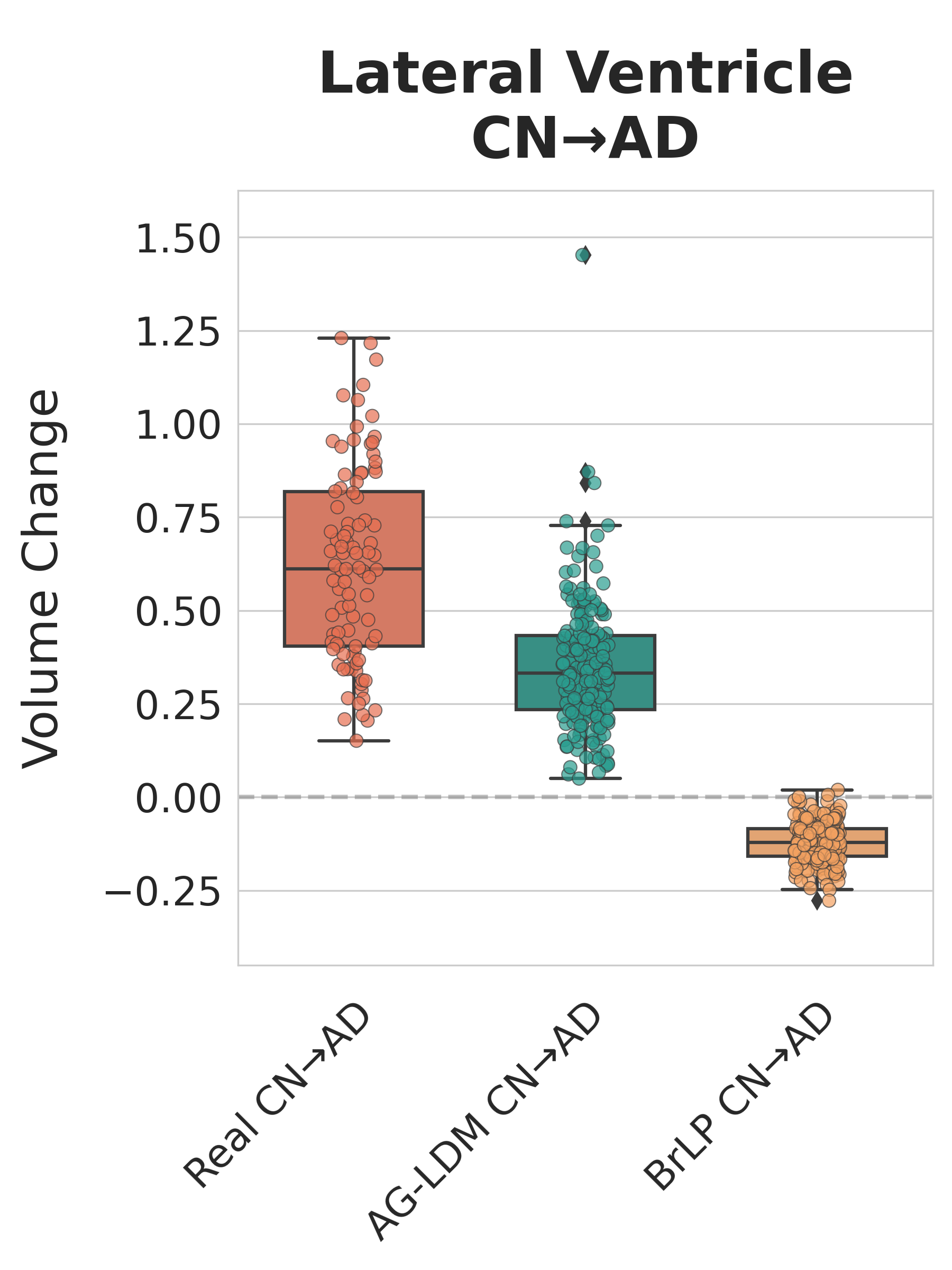}
\end{subfigure}\hfill
\begin{subfigure}[t]{\sixw}\centering
  \includegraphics[width=\linewidth]{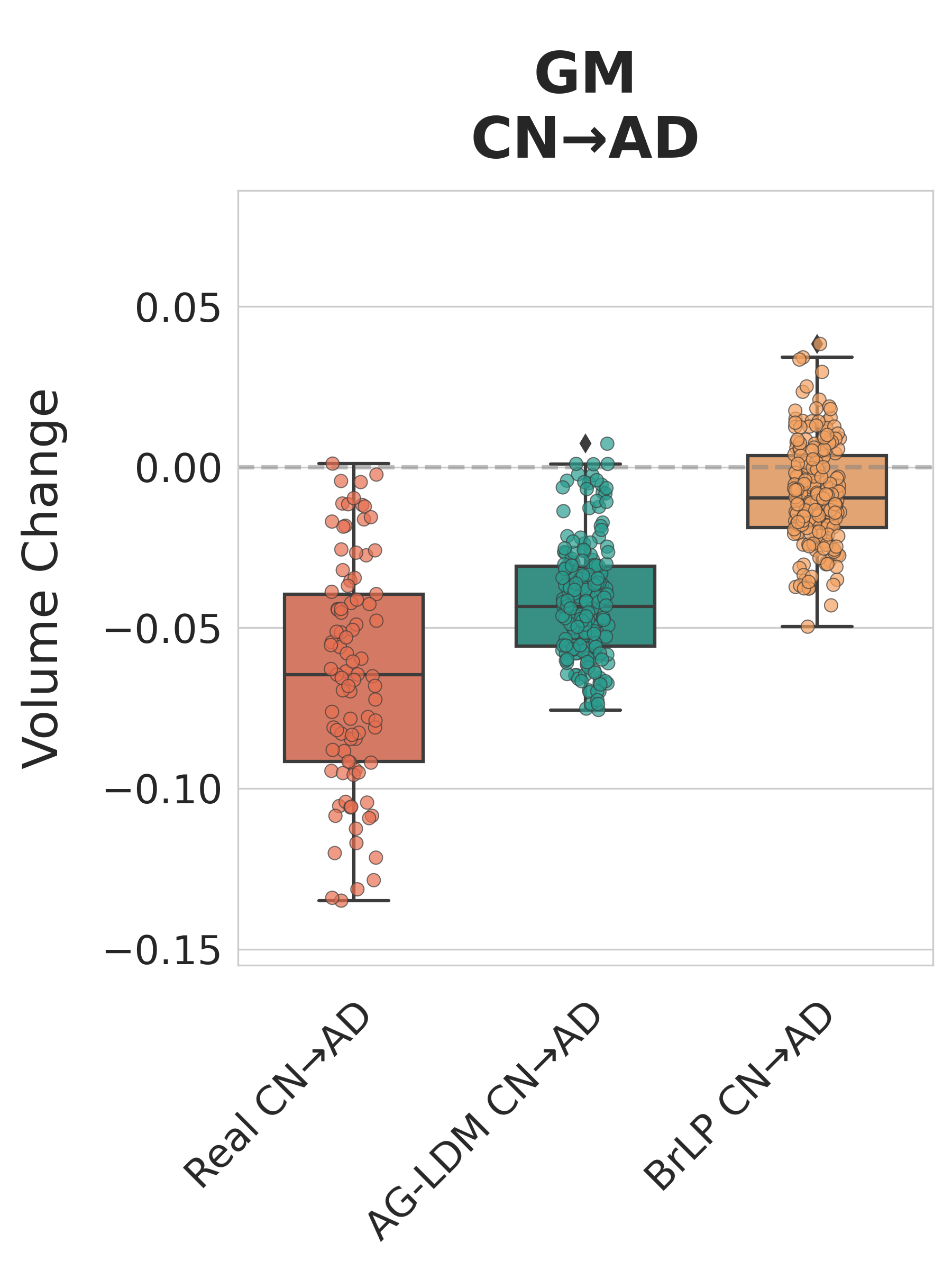}
\end{subfigure}\hfill
\begin{subfigure}[t]{\sixw}\centering
  \includegraphics[width=\linewidth]{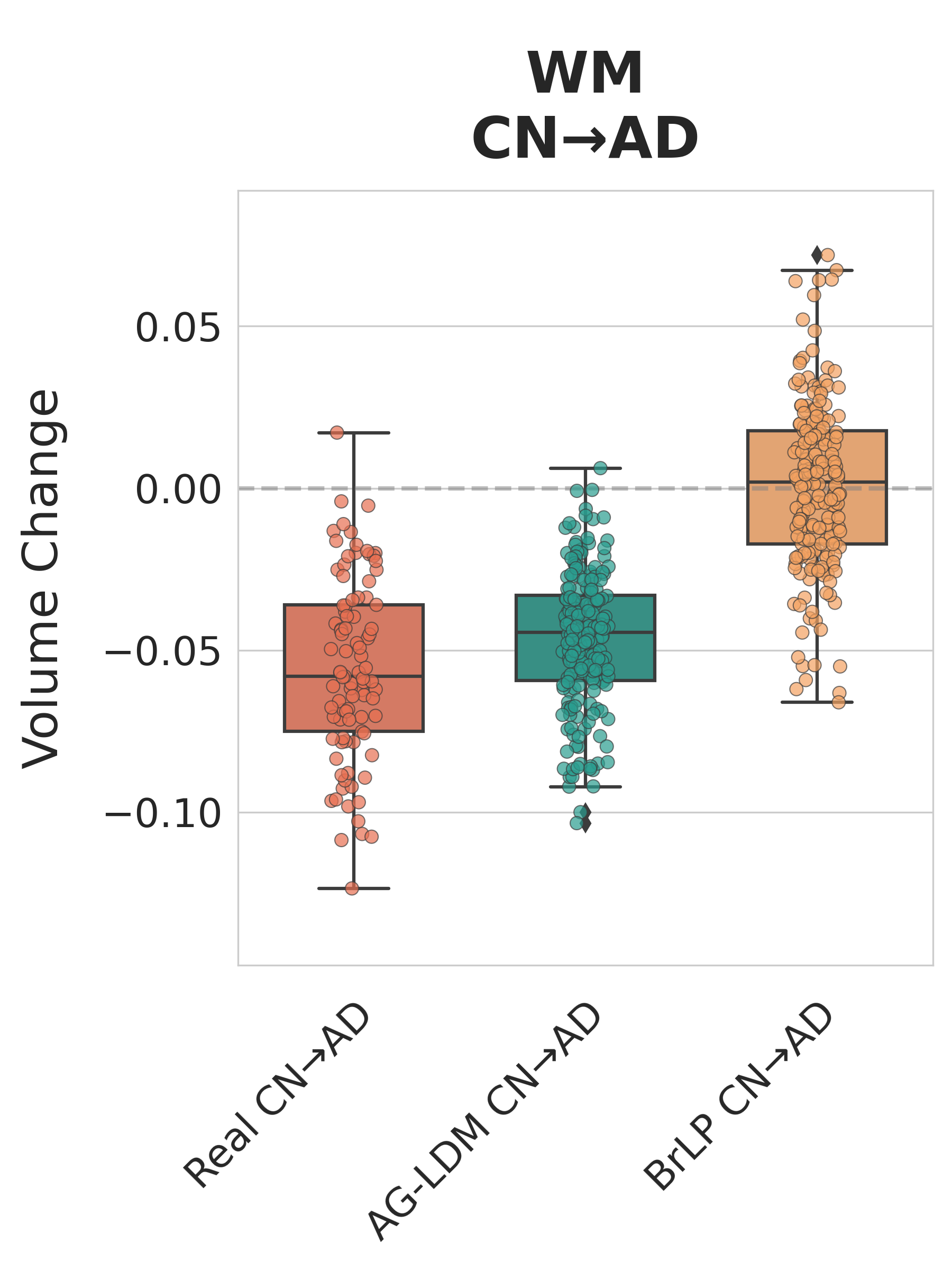}
\end{subfigure}

\vspace{2pt}

% ---------- Row 2: CN -> CN ----------
\begin{subfigure}[t]{\sixw}\centering
  \includegraphics[width=\linewidth]{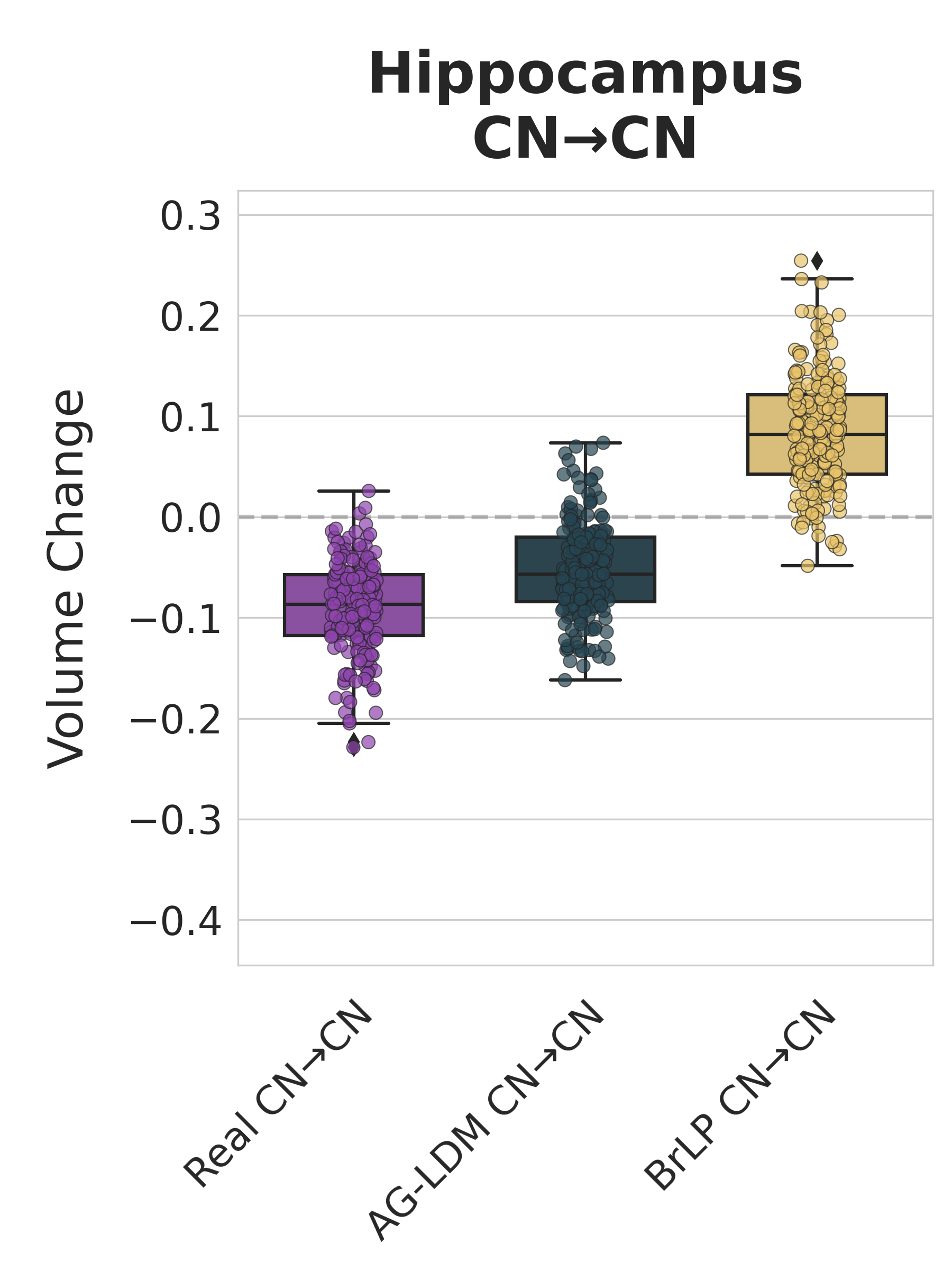}
\end{subfigure}\hfill
\begin{subfigure}[t]{\sixw}\centering
  \includegraphics[width=\linewidth]{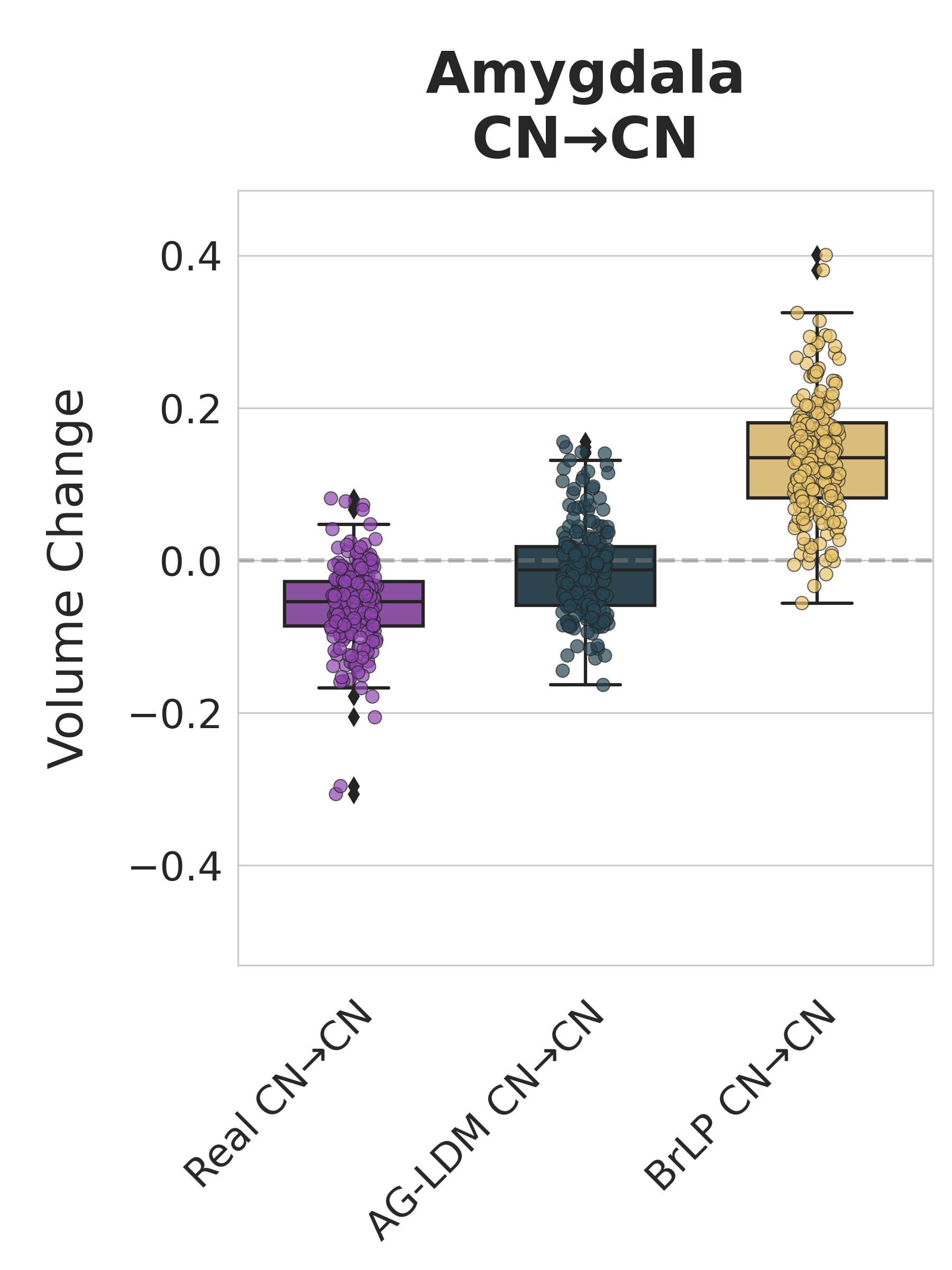}
\end{subfigure}\hfill
\begin{subfigure}[t]{\sixw}\centering
  \includegraphics[width=\linewidth]{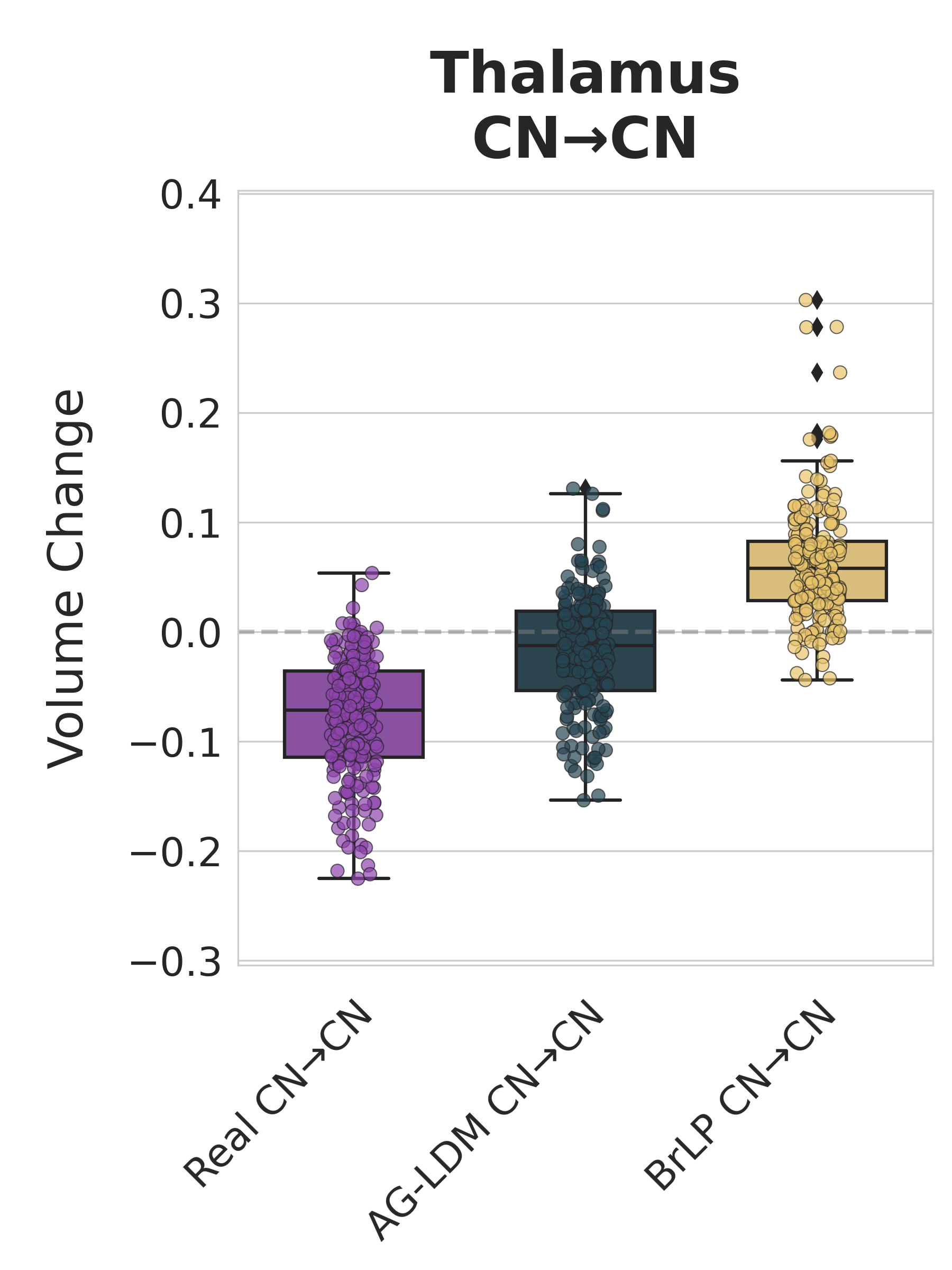}
\end{subfigure}\hfill
\begin{subfigure}[t]{\sixw}\centering
  \includegraphics[width=\linewidth]{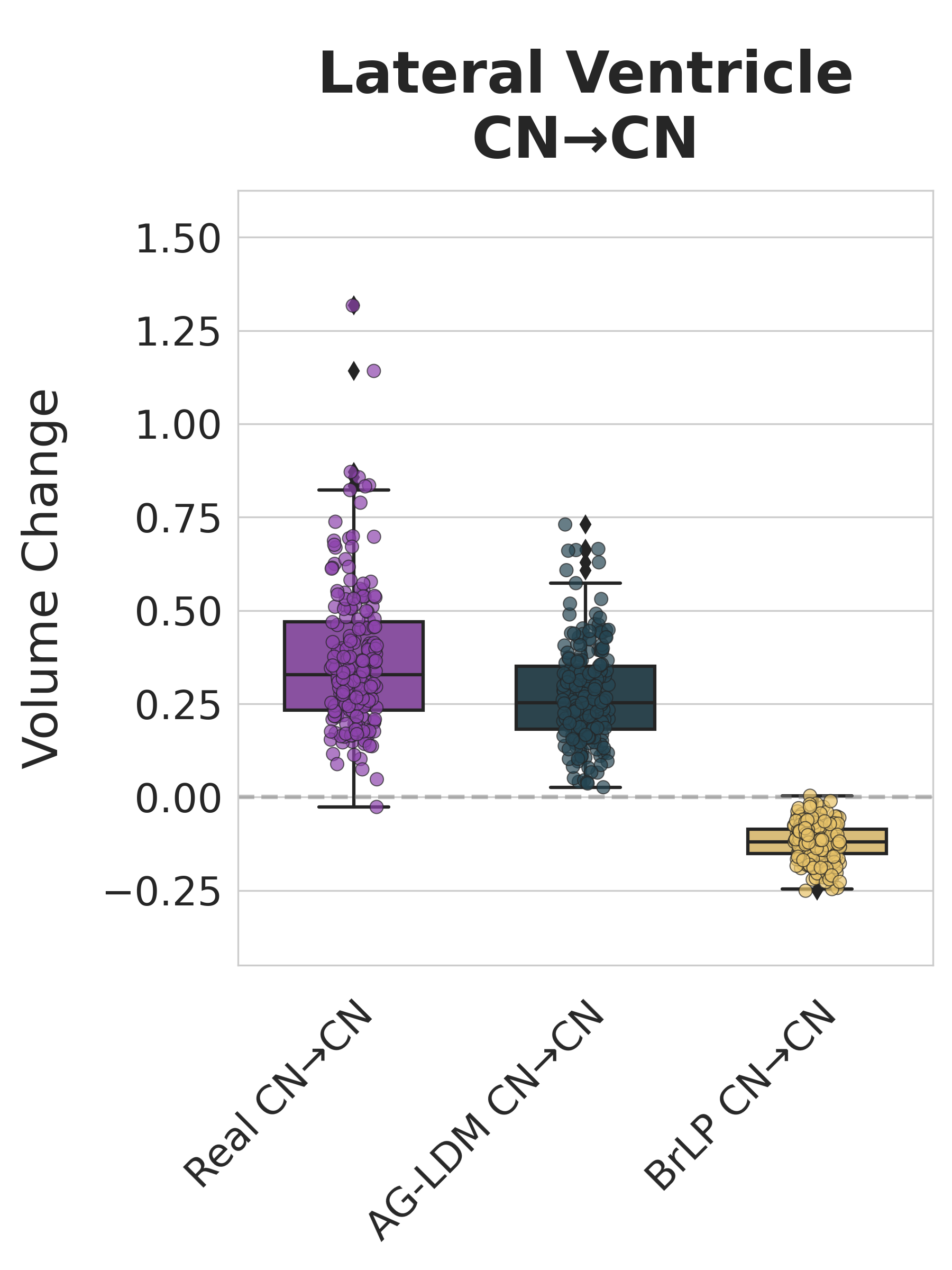}
\end{subfigure}\hfill
\begin{subfigure}[t]{\sixw}\centering
  \includegraphics[width=\linewidth]{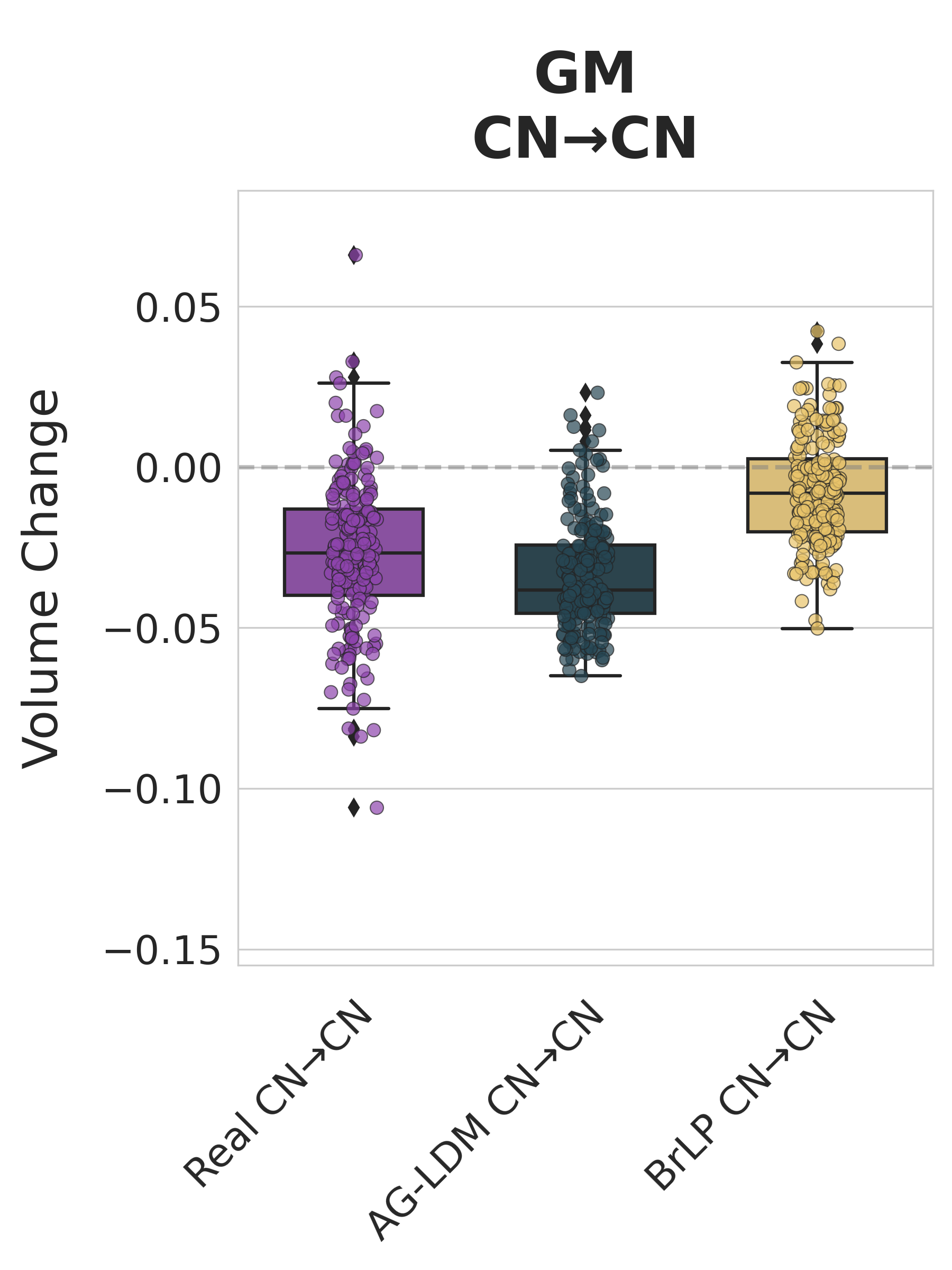}
\end{subfigure}\hfill
\begin{subfigure}[t]{\sixw}\centering
  \includegraphics[width=\linewidth]{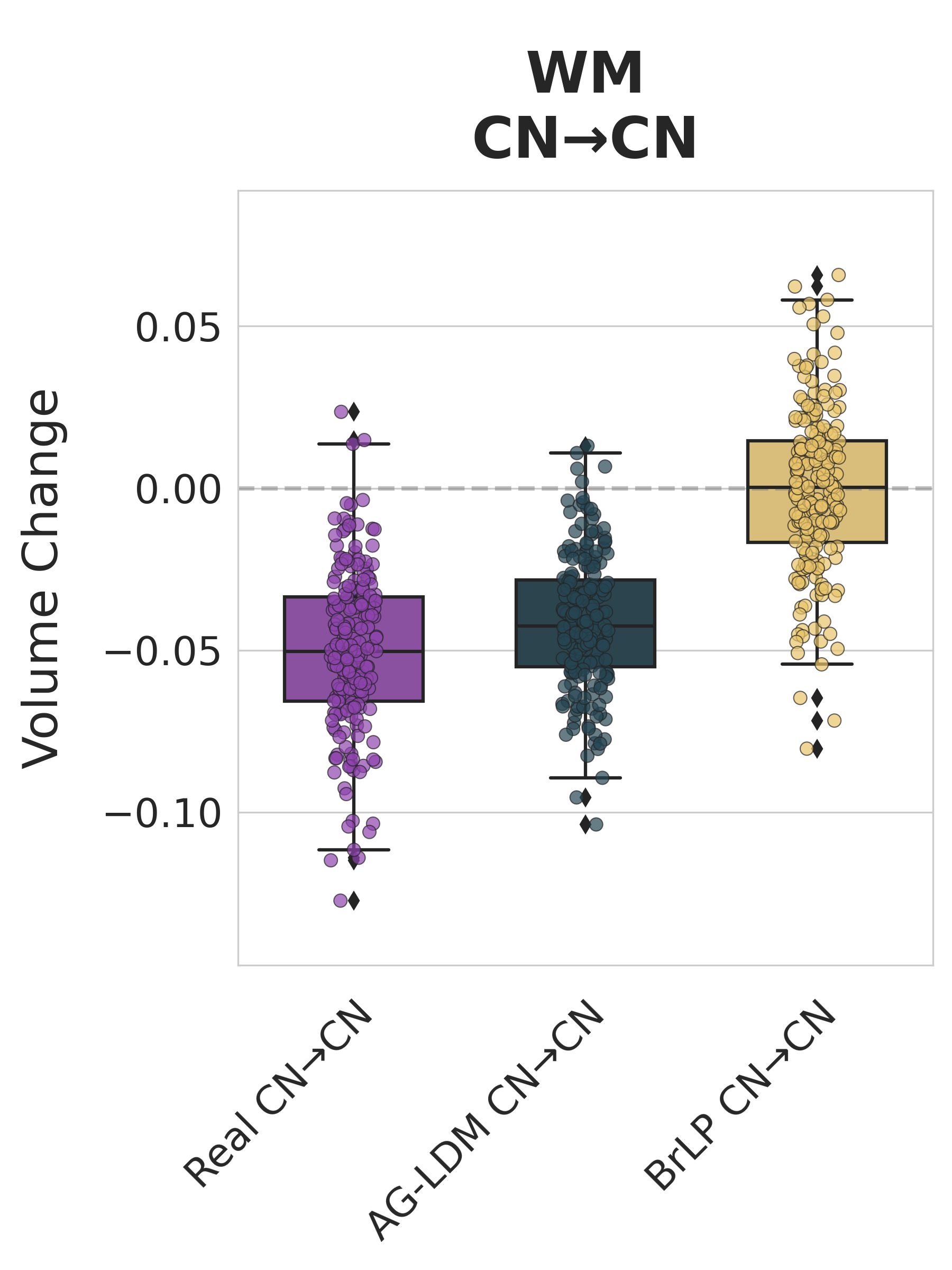}
\end{subfigure}

\caption{\textbf{Comparing Real and Synthesized Trajectories:} Top row: comparing observed volume change among 93 real subjects converting from CN to AD with trajectories counterfactually synthesized by AG-LDM and BrLP. Bottom row: comparing observed volume change among 200 real CN subjects with trajectories synthesized by AG-LDM and BrLP.
Each tile shows the relative volume change ($\Delta V_{\text{rel}}$) in an ROI between the baseline and (real or synthesized) follow-up MRIs.}
\label{fig:cf_cn_rowsix}
\end{figure*}
We assess whether the model captures normal aging and accelerated aging linked to AD through a counterfactual experiment on the ADNI test set.
Specifically, we identify 200 pairs of longitudinal scans from subjects who remained CN at both timepoints, constraining that each pair has a time interval greater than 7 years to evaluate the model's capacity for long-term progression synthesis.
Using the baseline scan and the \emph{actual follow-up age} as inputs, we use AG-LDM to synthesize two future scans for each subject: 
1) a \emph{natural} trajectory conditioned on the diagnosis remaining CN (AG-LDM CN$\rightarrow$CN), and 
2) a \emph{counterfactual} trajectory conditioned on the diagnosis converting to AD (AG-LDM CN$\rightarrow$AD). 
For comparison, we also simulate the two future scans by BrLP.
These synthesized trajectories are compared against the ground truth real longitudinal changes of the same 200 subjects (Real CN$\rightarrow$CN) and, for reference, against a separate group of real ADNI subjects who converted from CN to AD (Real CN$\rightarrow$AD, $n{=}93$). 

We analyze real and synthesized trajectories of hippocampus, amygdala, thalamus, lateral ventricles, GM, and WM. All synthesized and real follow-up images are segmented with SynthSeg.
For each subject and ROI, we compute head-size-normalized volumes and define the relative volume change as $\Delta V_{\text{rel}} = (V_{\text{follow-up}}-V_{\text{baseline}})/V_{\text{baseline}}\times 100\%$, where a positive $\Delta V_{\text{rel}}$ indicates volume expansion and a negative value indicates atrophy. Outliers are removed using an IQR rule. Figure~\ref{fig:cf_cn_rowsix} summarizes the distribution of $\Delta V_{\text{rel}}$ across real data, AG-LDM simulations, and BrLP simulations.

The real data show expected biological patterns: mild atrophy and moderate ventricular expansion during normal aging, versus sharper AD-related decline, especially in limbic regions (hippocampus and amygdala), with pronounced ventricular enlargement. AG-LDM approximates these signatures more closely than BrLP, aligning with the directionality of progression and preserving the relative ordering of effect sizes across ROIs (e.g., showing greater sensitivity in the hippocampus compared to the thalamus).
For instance, under the counterfactual AD condition, AG-LDM correctly predicts hippocampal atrophy ($-7.3 \pm 5.0\%$), whereas BrLP erroneously predicts growth ($+8.1 \pm 5.4\%$). 
Similarly, for the \textbf{lateral ventricles}, while real AD subjects and AG-LDM's counterfactual predictions show massive expansion ($+62.0 \pm 26.1\%$ and $+34.8 \pm 17.1\%$), BrLP predicts shrinkage ($-12.1 \pm 5.4\%$). 
Furthermore, AG-LDM successfully differentiates disease severity (e.g., hippocampal atrophy increases from $-5.2\%$ in natural aging to $-7.3\%$ in counterfactual AD), whereas BrLP produces nearly identical erroneous trajectories across conditions.

These results suggest that the conditioning mechanism in AG-LDM effectively steers synthesis along biologically plausible trajectories. The improved alignment between AG-LDM's counterfactual simulations and observed longitudinal trends supports its potential utility for simulation and hypothesis testing in studies of brain aging and neurodegeneration.

\section{Conclusion} In this work, we presented AG-LDM, a streamlined latent diffusion framework for brain MRI progression modeling that prioritizes anatomical consistency through explicit segmentation guidance. By simplifying the conditioning mechanism and integrating a lightweight segmentation teacher, AG-LDM achieves highly competitive image quality and significantly reduces volumetric errors compared to complex multi-stage architectures. Our experiments demonstrate that it effectively leverages clinical covariates to generate biologically plausible counterfactual trajectories, offering a reliable tool for visualizing individualized disease progression. 
A limitation of the current framework is reliance on a single frozen segmentation teacher (WarpSeg) during training. Although evaluation uses an independent tool (SynthSeg) on both the internal ADNI test set and zero-shot OASIS-3 cohort, with consistent results across cohorts, a systematic study of how the choice of segmentation teacher (e.g., WarpSeg vs.\ SynthSeg- or FastSurfer-distilled alternatives) influences the learned generator remains an important direction for future work.
Furthermore, while the current framework focuses on predicting specific follow-up scans, future work will extend AG-LDM to full longitudinal modeling, capturing complete temporal trajectories across multiple time points to provide a continuous and comprehensive view of neurodegenerative evolution.
% \section*{Acknowledgment}
% The authors thank the ADNI and OASIS consortiums for providing access to longitudinal brain imaging data. We also acknowledge computational support from the High Performance Computing Center.

\bibliographystyle{abbrv}
\bibliography{refs}

\clearpage
\appendix
\setcounter{figure}{0}
\setcounter{table}{0}
\renewcommand{\thefigure}{S\arabic{figure}}
\renewcommand{\thetable}{S\arabic{table}}
\renewcommand{\figurename}{Supplementary Figure}
\renewcommand{\tablename}{Supplementary Table}
\section{Network Architecture and Sampling}
\label{app:arch}

The Stage-2 noise-prediction network $\boldsymbol{\epsilon}_\theta$ is a 3D U-Net (MONAI \texttt{DiffusionModelUNet}) that operates entirely in the compressed latent space of the frozen autoencoder. At each diffusion timestep $t$, its input is the $11$-channel tensor of Eq.~\eqref{eq:concat}, formed by concatenating the noisy follow-up latent $\mathbf{z}_t^{(B)}$ ($3$ channels), the clean baseline latent $\mathbf{z}^{(A)}$ ($3$ channels), and the spatially broadcast clinical-covariate tensor $\mathbf{c}$ ($5$ channels). All conditioning is therefore fused at the input layer, so the same backbone represents both anatomy and progression without a separate control branch. The network uses three resolution levels with channel widths $(256, 512, 768)$, two residual blocks per level, self-attention at the two coarsest resolutions, GroupNorm ($32$ groups) with SiLU activations, and a sinusoidal timestep embedding that is projected by a two-layer MLP and added to every residual block. Supplementary Table~\ref{tab:supp_unet} lists the full layer-level specification.

\begin{table}[htbp]
\centering
\small
\renewcommand{\arraystretch}{1.2}
\caption{Layer-level specification of the AG-LDM Stage-2 diffusion U-Net $\boldsymbol{\epsilon}_\theta$ (MONAI \texttt{DiffusionModelUNet}, $475.4$~M trainable parameters). The 11-channel input is the concatenation $[\mathbf{z}_t^{(B)} \,(3)\,|\, \mathbf{z}^{(A)} \,(3)\,|\, \mathbf{c}\,(5)]$. Spatial resolution is given for a $122 \times 146 \times 122$ input volume (latent $15 \times 18 \times 15$).}
\label{tab:supp_unet}
\begin{tabular}{@{}l l c c c c@{}}
\toprule
\textbf{Stage} & \textbf{Block} & \textbf{Channels (in$\to$out)} & \textbf{Spatial Res.} & \textbf{Res.\ Blocks} & \textbf{Self-Attn} \\
\midrule
Input  & Conv in              & $11 \to 256$                  & $15\!\times\!18\!\times\!15$ & --  & no  \\
Level 0 & Residual$\times$2   & $256 \to 256$                 & $15\!\times\!18\!\times\!15$ & 2   & no  \\
Down    & Downsample          & $256 \to 256$                 & $8\!\times\!9\!\times\!8$    & --  & --  \\
Level 1 & Residual$\times$2 + SA & $256 \to 512$              & $8\!\times\!9\!\times\!8$    & 2   & yes \\
Down    & Downsample          & $512 \to 512$                 & $4\!\times\!5\!\times\!4$    & --  & --  \\
Level 2 & Residual$\times$2 + SA & $512 \to 768$              & $4\!\times\!5\!\times\!4$    & 2   & yes \\
\midrule
\multicolumn{6}{l}{\textit{(mirrored decoder with skip connections from the encoder at each level)}} \\
\midrule
Output  & Conv out             & $256 \to 3$                   & $15\!\times\!18\!\times\!15$ & --  & no  \\
\bottomrule
\multicolumn{6}{l}{Time embedding: 1024-d sinusoidal $\to$ Linear-SiLU-Linear MLP $\to$ added to every residual block.} \\
\multicolumn{6}{l}{Normalization: GroupNorm (32 groups) throughout. Activation: SiLU.} \\
\end{tabular}
\end{table}

\paragraph{Deterministic sampling.}
Training minimizes the noise-prediction error under a DDPM process~\cite{ho2020denoising} with $T=1000$ steps and a scaled-linear schedule. At inference, and for the differentiable anatomical supervision below, we use deterministic DDIM sampling~\cite{song2020denoising}. Given the network output $\boldsymbol{\epsilon}_\theta(\mathbf{z}_t,t)$, the estimated clean latent and the previous state are
\begin{equation}
\tilde{\mathbf{z}}_0 = \frac{\mathbf{z}_t - \sqrt{1-\bar\alpha_t}\;\boldsymbol{\epsilon}_\theta(\mathbf{z}_t,t)}{\sqrt{\bar\alpha_t}},
\qquad
\mathbf{z}_{t-1} = \sqrt{\bar\alpha_{t-1}}\;\tilde{\mathbf{z}}_0 + \sqrt{1-\bar\alpha_{t-1}}\;\boldsymbol{\epsilon}_\theta(\mathbf{z}_t,t),
\label{eq:ddim}
\end{equation}
where $\bar\alpha_t=\prod_{i=1}^{t}\alpha_i$ follows the same schedule as the forward process. We run $50$ DDIM steps at test time. During training, the segmentation-guided term of Eq.~\eqref{eq:ldm_total} uses a shallow $10$-step DDIM unroll to obtain $\tilde{\mathbf{z}}_0^{(B)}$; since the decoder $\mathcal{D}$ and the frozen segmentation teacher $\Phi$ are differentiable, gradients of the Dice term propagate through $\Phi\!\to\!\mathcal{D}\!\to\!\boldsymbol{\epsilon}_\theta$ across all $10$ unrolled steps, so $\boldsymbol{\epsilon}_\theta$ is the only updated module.

% =====================================================================
\section{Autoencoder Reconstruction Quality}
\label{app:aerecon}

\begin{wrapfigure}{r}{0.55\textwidth}
\centering
\vspace{-\intextsep}
\includegraphics[width=0.53\textwidth]{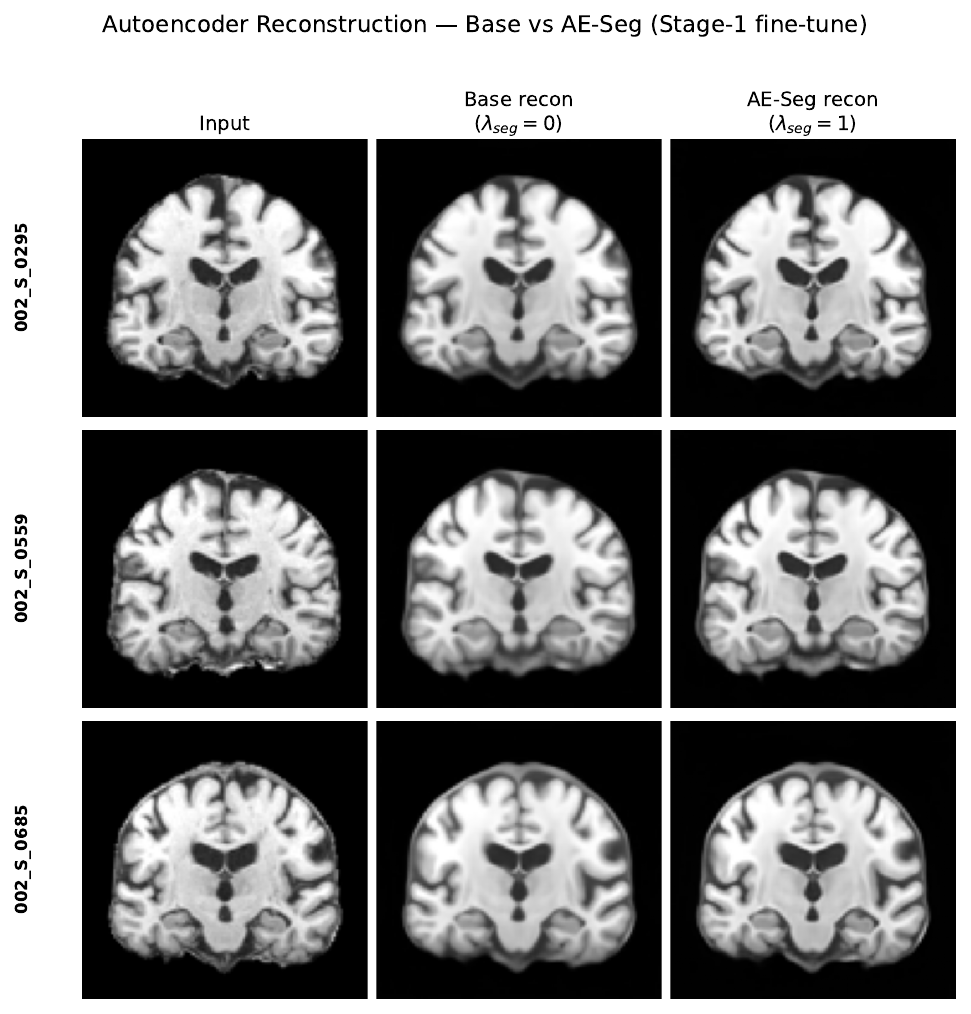}
\caption{Qualitative AE reconstruction on three held-out ADNI subjects. \textbf{Left}: AE input. \textbf{Middle}: baseline AE ($\lambda_{\text{seg}}=0$). \textbf{Right}: AE-Seg ($\lambda_{\text{seg}}=1$). AE-Seg yields sharper tissue boundaries.}
\label{fig:supp_recon}
\end{wrapfigure}

Stage-1 fine-tunes the pre-trained autoencoder with the tissue-segmentation objective $\mathcal{L}_{\text{seg}}$ of Eq.~\eqref{eq:seg_loss}, using a lightweight frozen teacher (WarpSeg) to supervise gray- and white-matter boundaries. Because this supervision acts on the reconstructed image, it constrains the learned latent space to preserve tissue geometry in addition to intensity fidelity.

To isolate the contribution of this stage, we compare the baseline autoencoder (Base, $\lambda_{\text{seg}}=0$) against the segmentation-guided variant (AE-Seg, $\lambda_{\text{seg}}=1$) on the self-reconstruction of held-out ADNI baseline scans, repeated across the eight matched hyperparameter configurations used throughout the ablation. Quantitatively, AE-Seg improves every tissue-level reconstruction metric (Dice and volume MAE) reported in the main text, with the largest relative gains on cerebrospinal fluid, whose thin structures are most sensitive to boundary blurring. Supplementary Figure~\ref{fig:supp_recon} shows representative examples: relative to Base, AE-Seg reconstructions recover gray--white matter interfaces more crisply and reduce the smoothing that a purely intensity-based objective tends to introduce. These qualitative differences are consistent with the numerical improvements and indicate that anatomical supervision at Stage-1 yields a latent representation better suited to the downstream progression task.

% =====================================================================
\section{Segmentation Guidance at the Diffusion Stage}
\label{app:stage2}

\begin{wraptable}{r}{0.55\textwidth}
\centering
\small
\renewcommand{\arraystretch}{1.2}
\caption{Tissue-level Dice of the \emph{generated} follow-up scans for AE-Seg and Full, averaged over the eight hyperparameter configurations (mean $\pm$ SD across the per-configuration validation-set means). $^{*}$ Full significantly outperforms AE-Seg (paired one-sided $t$-test, $p<0.001$).}
\label{tab:supp_stage2}
\resizebox{0.52\textwidth}{!}{%
\begin{tabular}{l c c c}
\toprule
\textbf{Tissue} & \textbf{AE-Seg} & \textbf{Full} & \textbf{$\Delta$} \\
\midrule
GM Dice  & $0.800 \pm 0.054$ & $\mathbf{0.810 \pm 0.051}^{*}$ & $+0.010$ \\
WM Dice  & $0.865 \pm 0.034$ & $\mathbf{0.873 \pm 0.031}^{*}$ & $+0.009$ \\
CSF Dice & $0.728 \pm 0.103$ & $\mathbf{0.745 \pm 0.101}^{*}$ & $+0.017$ \\
\bottomrule
\end{tabular}}
\end{wraptable}

Anatomical supervision is applied not only during autoencoder fine-tuning but also during diffusion training, through the Dice term of the Stage-2 objective in Eq.~\eqref{eq:ldm_total} (the Full model). To characterize the marginal contribution of this second stage, we compare AE-Seg and Full on the tissue-level Dice of the \emph{generated} follow-up scans, averaged over the eight matched hyperparameter configurations.

As summarized in Supplementary Table~\ref{tab:supp_stage2}, the diffusion-stage supervision provides a small but statistically consistent improvement over AE-Seg for all three tissue types (paired one-sided $t$-test over the eight configurations, $p<0.001$). The gain is modest in magnitude, which is expected since much of the anatomical benefit is already secured during Stage-1; its role is to further tighten the tissue boundaries of the synthesized follow-up scan, rather than of the reconstruction. Consistent with this, the segmentation-guided variants also exhibit narrower cross-configuration Dice distributions, indicating that anatomical supervision additionally stabilizes training against the choice of Stage-1 hyperparameters.

% =====================================================================
\section{Sensitivity to the Segmentation-Loss Weight \texorpdfstring{$\gamma$}{γ}}
\label{app:gamma}

\begin{wraptable}{r}{0.55\textwidth}
\centering
\small
\renewcommand{\arraystretch}{1.15}
\caption{Sensitivity of the Full model to the Dice-loss weight $\gamma$, with the supervision applied at every iteration. Each row is mean $\pm$ SD over the ADNI validation set; the value $\gamma=10^{-5}$ used in the main experiments is shown in bold.}
\label{tab:supp_gamma}
\resizebox{0.52\textwidth}{!}{%
\begin{tabular}{l cccc}
\toprule
\textbf{$\gamma$} & \textbf{MSE} ($\times 10^{-3}$) $\downarrow$ & \textbf{PSNR} $\uparrow$ & \textbf{WM Dice} $\uparrow$ & \textbf{GM Dice} $\uparrow$ \\
\midrule
$10^{-7}$ & $4.04 \pm 2.40$ & $24.51 \pm 2.13$ & $0.887 \pm 0.030$ & $0.831 \pm 0.041$ \\
$10^{-6}$ & $3.85 \pm 2.35$ & $24.73 \pm 2.15$ & $0.890 \pm 0.029$ & $0.836 \pm 0.040$ \\
$\mathbf{10^{-5}}$ & $\mathbf{3.81 \pm 2.14}$ & $24.72 \pm 2.08$ & $\mathbf{0.891 \pm 0.028}$ & $\mathbf{0.837 \pm 0.039}$ \\
$10^{-4}$ & $3.91 \pm 2.38$ & $24.66 \pm 2.14$ & $0.889 \pm 0.029$ & $0.835 \pm 0.040$ \\
$10^{-3}$ & $4.07 \pm 2.42$ & $24.47 \pm 2.12$ & $0.887 \pm 0.030$ & $0.832 \pm 0.041$ \\
\bottomrule
\end{tabular}}
\end{wraptable}

The weight $\gamma$ in the Stage-2 objective of Eq.~\eqref{eq:ldm_total} balances the noise-prediction loss $\mathcal{L}_{\text{noise}}$ against the anatomical Dice term $\mathcal{L}_{\text{dice}}$. It is the only hyperparameter that couples the two objectives, and its value determines whether anatomical supervision behaves as a gentle regularizer or as a competing loss; we therefore examine its influence directly.

We retrain the Full model with $\gamma \in \{10^{-7}, 10^{-6}, 10^{-5}, 10^{-4}, 10^{-3}\}$, keeping every other setting fixed and applying the supervision at each iteration, and evaluate on the ADNI validation set. As shown in Supplementary Table~\ref{tab:supp_gamma}, image quality (MSE, PSNR) and tissue Dice remain within a narrow band across four orders of magnitude, $\gamma \in [10^{-7}, 10^{-4}]$: the MSE varies by under $\sim$7\% and the tissue Dice by less than $0.01$ over this range. Performance degrades only at the largest value, $\gamma=10^{-3}$. The value $\gamma=10^{-5}$ used in all main experiments lies in the centre of this stable region and attains the best MSE and WM/GM Dice in the sweep.

\paragraph{Gradient-balance analysis.}
This stability across several orders of magnitude of $\gamma$ can be understood from the relative scale of the two gradient contributions. At every step the parameters follow
\begin{equation}
\nabla_\theta \mathcal{L}_{\text{LDM}} = \nabla_\theta \mathcal{L}_{\text{noise}} + \gamma\,\nabla_\theta \mathcal{L}_{\text{dice}},
\end{equation}
and we define the relative magnitude of the anatomical term as
\begin{equation}
r(\gamma) \;=\; \frac{\gamma\,\lVert \nabla_\theta \mathcal{L}_{\text{dice}}\rVert}{\lVert \nabla_\theta \mathcal{L}_{\text{noise}}\rVert}.
\label{eq:gamma_ratio}
\end{equation}
Empirically, the raw ratio $\lVert \nabla_\theta \mathcal{L}_{\text{dice}}\rVert / \lVert \nabla_\theta \mathcal{L}_{\text{noise}}\rVert$ stays roughly constant over training, so $r(\gamma)$ grows approximately linearly with $\gamma$; at the selected $\gamma=10^{-5}$ the anatomical contribution is about an order of magnitude below the noise-prediction gradient ($r\approx 0.1$) and reaches parity ($r\approx 1$) only near $\gamma=10^{-4}$.

When $r(\gamma)\ll 1$, the dependence of the trained model on $\gamma$ admits a first-order characterization. Let $\theta^\star(\gamma)$ be a stationary point of the objective, satisfying $\nabla_\theta \mathcal{L}_{\text{noise}}(\theta^\star) + \gamma\,\nabla_\theta \mathcal{L}_{\text{dice}}(\theta^\star) = \mathbf{0}$, and let $\theta^\star_0$ be the noise-only solution at $\gamma=0$. Assuming $\theta^\star_0$ is a local minimum of $\mathcal{L}_{\text{noise}}$ with non-singular Hessian $\mathbf{H}=\nabla^2_\theta \mathcal{L}_{\text{noise}}(\theta^\star_0)$, the implicit function theorem yields the first-order expansion
\begin{equation}
\theta^\star(\gamma) \;\approx\; \theta^\star_0 \;-\; \gamma\,\mathbf{H}^{-1}\,\nabla_\theta \mathcal{L}_{\text{dice}}(\theta^\star_0),
\label{eq:gamma_expansion}
\end{equation}
so the induced parameter shift is $\mathcal{O}(\gamma)$ and any smooth evaluation metric $M$ obeys $M(\theta^\star(\gamma)) = M(\theta^\star_0) + \mathcal{O}(\gamma)$. Over $\gamma\in[10^{-7},10^{-4}]$ this predicts a small, gradual drift rather than a regime change, in agreement with the narrow band of Supplementary Table~\ref{tab:supp_gamma}. The approximation ceases to hold once $r(\gamma)$ approaches one---near $\gamma=10^{-3}$ in our setting---where the anatomical gradient is no longer a perturbation and begins to trade off against accurate noise prediction; this is precisely the regime in which the measured image quality degrades. We emphasize that this is a local, first-order argument that rationalizes the observed insensitivity within the perturbative regime, not a global guarantee.

\end{document}